%% file: root.tex
\documentclass[journal]{IEEEtran}  

\IEEEoverridecommandlockouts                              

\usepackage{graphicx}
\usepackage{cite}
\usepackage[linesnumbered,ruled,vlined]{algorithm2e}
\usepackage[hidelinks]{hyperref}
\usepackage{amsmath}
\usepackage{amssymb}
\usepackage{booktabs}
\usepackage{multirow}
\usepackage{multicol}
\usepackage{makecell}
\usepackage{anyfontsize}
\usepackage{capt-of}
\usepackage{siunitx}
\usepackage{xspace}
\newcommand{\po}{\texttt{Place Orange}\xspace}
\newcommand{\fe}{\texttt{Flip Egg}\xspace}
\newcommand{\ob}{\texttt{Open Box}\xspace}
\newcommand{\ib}{\texttt{Insert Bills}\xspace}
\usepackage{bbm}
\usepackage{stfloats}

\newcommand{\ie}{i.e.,\xspace}

\usepackage{mathtools}
\title{
   Beyond Action Residuals: Real-World Robot Policy Steering via Bottleneck Latent Reinforcement Learning
}

\author{
Dongjie Yu$^{1,2,*}$, Kun Lei$^{2,3,*}$, Zhennan Jiang$^{4}$, Jia Pan$^{1,\dagger}$, and Huazhe Xu$^{2,5,\dagger}$%
\thanks{$^{*}$Equal contribution. $^{\dagger}$Corresponding authors: Jia Pan and Huazhe Xu.}%
\thanks{$^{1}$School of Computing and Data Science, The University of Hong Kong, Hong Kong SAR, China (e-mail: djyu@connect.hku.hk; jpan@cs.hku.hk).}%
\thanks{$^{2}$Shanghai Qizhi Institute, Shanghai, China.}%
\thanks{$^{3}$Shanghai Jiao Tong University, Shanghai, China (e-mail: leikun980116@gmail.com).}%
\thanks{$^{4}$Institute of Automation, Chinese Academy of Sciences, Beijing, China (e-mail: jiangzhennan2024@ia.ac.cn).}%
\thanks{$^{5}$Institute for Interdisciplinary Information Sciences, Tsinghua University, Beijing, China (e-mail: huazhe\_xu@mail.tsinghua.edu.cn).}%
}

\input{math_symbols}

\begin{document}

\bstctlcite{BSTcontrol}

\maketitle

\begin{abstract}
Pretrained imitation policies have become a strong foundation for robot manipulation, but online adaptation can further improve policy performance within their pretrained behavioral support. A central question is therefore how reinforcement learning (RL) should adapt policies after offline pretraining.
Existing lightweight methods commonly apply residual corrections directly in action space. We investigate how the steering interface affects online RL. In this work, we propose Z-Perturbation Reinforcement Learning (ZPRL), an approach that steers pretrained policies through a compact bottleneck latent rather than through policy weights or output actions.
Building on an offline-learned bottleneck policy interface introduced by previous work, ZPRL turns this representation into an online RL control space through reward-driven residual perturbations while freezing the pretrained policy. We instantiate ZPRL on flow-matching policies and evaluate it on eight simulation tasks and four real-world tasks. Across diverse manipulation settings, ZPRL improves sample efficiency with competitive or better final performance.
In the real world, ZPRL improves the average success rate on four tasks by 33.7 percentage points over imitation base policies. Our qualitative and quantitative results also indicate that ZPRL combines task success with smooth actions without explicit temporal filtering or smoothness regularization. These results suggest that a compact, task-aligned bottleneck latent provides an effective interface for online RL adaptation. Videos and code can be found at https://manutdmoon.github.io/ZPRL/.
\end{abstract}


\section{Introduction}

Imitation learning (IL) on offline datasets has become an increasingly popular approach to building robot manipulation policies given demonstrations~\cite{lbm2025careful}. Recent policy classes, including transformers~\cite{black2025pi0,kim2025openvla} and generative sensorimotor policies such as variational auto-encoders~\cite{zhao2023aloha} or diffusion and flow models~\cite{ho2020ddpm,liu2023flow,chi2023diffusion,park2025flow,lu2026h3dp}, can reproduce complex behaviors from data with remarkable fidelity.
However, strong offline pretraining does not eliminate the need for online improvement: in real-world deployment, policies may possess the required behaviors yet fail due to errors in behavior selection or timing~\cite{luo2025precise}.
As a result, learning from interaction~\cite{jin2025sime,jin2025soe}, especially reinforcement learning (RL), remains an appealing mechanism for post-training robot policies, rather than relying exclusively on collecting more human data~\cite{luo2025precise,ren2025dppo,zhang2025reinflow,yuan2025hermes,lei2025rl100}.

A key question lies in how to apply RL to modern imitation-learned policies.
Full policy finetuning is increasingly costly as policy size grows~\cite{chen2026pirl,intelligence2025pi06}, and for generative policies it is often entangled with model-specific optimization designs~\cite{Shafiullah2022bet,ren2025dppo,zhang2025reinflow,yuan2025policy}.
To avoid these difficulties, recent lightweight adaptation methods freeze the base policy and learn only a corrective policy~\cite{Ankile2025from,ankile2025residual,Johannink2019residual,yuan2025policy}. This line of work is attractive because it preserves pretrained ability while reducing the burden of online adaptation.
However, when the correction is applied directly in action space, exploration remains low-level: the policy is encouraged to modify motor commands rather than the underlying action pattern~\cite{wagenmaker2025steering}. Without training-time regularization~\cite{yuan2025policy,mysore2021caps} or test-time curation, such action-space residuals may lead to oscillatory or erratic behavior, making exploration both inefficient and potentially unsafe.
Our intuition is that meaningful behaviors rarely emerge from unstructured oscillation; what is needed is not merely different actions, but different actions that have consistent pattern with the pretrained policy~\cite{jin2025soe}.

\begin{figure}[!t]
    \centering
    \includegraphics[width=0.95\columnwidth]{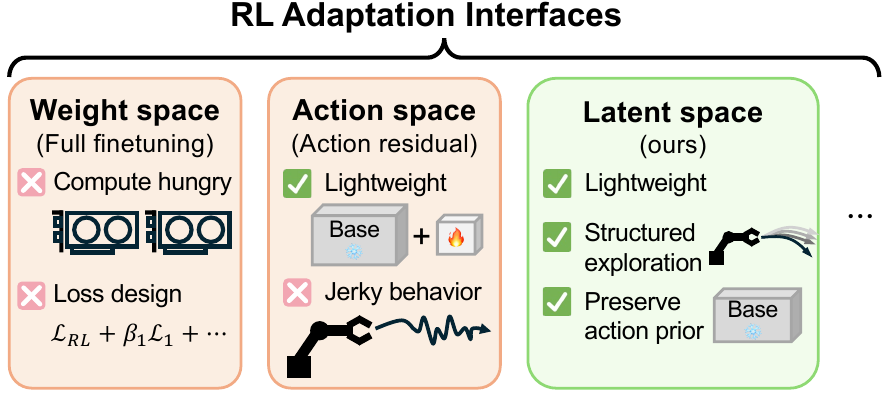}
    \vspace{-3mm}
    \caption{{\textbf{Different interfaces for RL adaptation of pretrained robot policies.} Full finetuning in weight space is expressive but computationally heavy and often tied to policy-specific loss designs. Residual adaptation in action space is lightweight, but exploration may be jerky and inefficient without special designs. ZPRL instead steers a compact bottleneck latent, providing a lightweight yet structured interface for online adaptation.}}
    \vspace{-5mm}
    \label{fig:motivation}
\end{figure}

We illustrate the design space of RL adaptation interfaces in~\Figref{fig:motivation}, which shows that the key issue is not only \emph{how much} of the policy should be adapted, but also \emph{where} RL should intervene.
A useful intervention space should be: (1) compact for efficient online steering, compared to full finetuning in weight space; and (2) structured to keep exploration near valid behaviors, compared to residual corrections in action space. Recent work has shown the promise of latent-space RL for diffusion policies~\cite{wagenmaker2025steering}, and has also shown that compact latent representations can support on-manifold exploration for bootstrap IL~\cite{jin2025soe}.
Therefore, a natural question emerges from these findings: instead of correcting actions directly, \textit{can RL steer a policy through a different but more efficient interface that better captures the structure of pretrained behaviors?}

In this work, we instantiate this idea as Z-Perturbation Reinforcement Learning (ZPRL). Following~\cite{alemi2017deep, jin2025soe}, ZPRL injects a plug-and-play variational information bottleneck (VIB) module at the observation-conditioning interface to extract a compact latent representation of task-relevant features during offline training.
Our contribution is not the VIB interface or latent steering itself: it is a residual-RL formulation that learns how to steer this interface from task reward within pretrained behavioral support, together with evidence that the choice of intervention space affects post-training efficiency and performance.
Compared with action-space residuals~\cite{Ankile2025from,ankile2025residual,Johannink2019residual,yuan2025policy}, this design moves exploration from low-level motor commands to structured latent steering. 
Compared with latent-space methods defined over diffusion noise~\cite{wagenmaker2025steering}, it uses a compact task-relevant bottleneck as the steering interface, aligning exploration with task-relevant features.

We evaluate ZPRL with both simulated manipulation benchmarks and real-world tasks, showing that it improves sample efficiency and final performance over baselines. After a few hours of real-world online interaction, ZPRL improves the average success rate (SR) by 33.7 percentage points (pp) over base policies.
Furthermore, ZPRL combines high task success with low acceleration and jerk without explicit temporal filtering or smoothness regularization. Our comparison with baselines evaluates this success-smoothness trade-off.
These results demonstrate that bottleneck latent steering provides an effective RL interface for post-training robot policies.

\section{Related Work}

\subsection{Imitation Learning in Robot Manipulation}

Robot manipulation with IL has evolved from low-dimensional skill representations, such as dynamical and probabilistic movement primitives~\cite{ijspeert2013dmp,paraschos2013promp,chu2026dammp}, to large-scale visuomotor behavior modeling with expressive neural architectures~\cite{zhao2023aloha, chi2023diffusion, ze2024dp3, intelligence2025pi06,tian2026vitas}. Benefiting from the rapid growth of robot datasets~\cite{zhu2020robosuite, Mandlekar2022robomimic, khazatsky2024droid, vuong2023open, li2023behavior} and compute resources, modern IL methods can build increasingly capable pretrained policies that perform a wide range of household and industrial manipulation tasks~\cite{lbm2025careful, black2025pi0}. 
Despite their impressive performance, these policies remain fundamentally constrained by the coverage and quality of offline data. In realistic deployments, task failures, execution errors, and distribution shifts often require further improvement beyond offline pretraining~\cite{du2025dynaguide}. This motivates methods that can adapt or steer pretrained policies through online interaction.

\subsection{Reinforcement Learning for Adapting Pretrained Policies}

Reinforcement learning provides a natural framework for improving pretrained policies beyond the limits of offline datasets~\cite{sutton1998reinforcement,levine2020offline}. A common approach is to finetune the entire policy end-to-end during online adaptation~\cite{lei2024unio,2026arXiv260107821L,lei2025rl100, uchendu2023jsrl, zhou2025wsrl, ball2023rlpd, luo2025precise, ren2025dppo, zhang2025reinflow}. From this perspective, adaptation occurs in the \emph{weight space} of the policy. While such methods can be effective for smaller models or policy classes with dedicated optimization procedures, full online finetuning becomes increasingly expensive for modern large-scale visuomotor and vision-language-action (VLA) policies, and often requires substantial systems and algorithmic support~\cite{ren2025dppo}.

A complementary line of work improves pretrained policies without updating the entire model. Residual RL methods perform adaptation in the \emph{action space} by learning corrective actions on top of classical controllers~\cite{silver2019residualpolicylearning,davchev2022residual} or frozen imitation-learned policies~\cite{Johannink2019residual, Ankile2025from, ankile2025residual, yuan2025policy,sun2026dicerl}. These methods are attractive because they preserve pretrained ability and provide a lightweight interface for online improvement. However, when the correction is applied directly in action space, exploration remains low-level, and can become inefficient in high-dimensional or chunked action spaces~\cite{li2025qchunk}. Their motion smoothness depends on the residual design, with temporal regularization offering a viable means of mitigating oscillation~\cite{mysore2021caps}. In addition, useful designs in residual RL~\cite{sun2026dicerl,ma2026what} can also be leveraged in our work because we inject residuals in a latent space.

More recent works have moved the RL intervention from action space to internal latent spaces, such as initial diffusion noise. These results highlight that the \emph{steering interface} itself is a critical design choice for RL post-training. Predicting independent noise for an entire action chunk increases the steering dimension with the action horizon, although repeating noise across time can reduce this dimension, as in DSRL (Diffusion Steering with RL)~\cite{wagenmaker2025steering}. In this work, we instead study a compact bottleneck latent as the steering interface, drawing inspiration from recent work on structured on-manifold exploration~\cite{jin2025soe}. Our method combines lightweight adaptation with a task-aligned low-dimensional latent space, enabling smooth behaviors and efficient online improvement.

\section{Preliminaries and Problem Formulation}

\subsection{Robot Manipulation as an Observation-Conditioned Decision Process}
We consider robot manipulation in an episodic decision process induced by an underlying Markov Decision Process (MDP) 
$\gM=\langle \gS, \gA, T, \rho_0, r, \gamma \rangle$,
where $\gS$ and $\gA$ denote the state and action spaces, respectively. In our setting, actions correspond to desired end-effector poses.
Since the policy does not access full environmental states, it instead acts on observations $\vo_t \in \gO$, which may include vision and robot proprioception.
At the beginning of each episode, the manipulation environment is initialized to $\vs_0 \sim \rho_0$, producing an initial observation $\vo_0$.
At each timestep $t$, the robot policy chooses an action $\va_t \sim \pi(\cdot | \vo_t)$, and then the environment transitions to the next state $\vs_{t+1} \sim T(\cdot | \vs_t, \va_t)$ and emits the next observation $\vo_{t+1}$. The policy receives a scalar reward $r_t = r(\vs_t, \va_t)$. Unless otherwise specified, we use sparse binary rewards, \ie $r_t=1$ upon task success (and the episode is terminated) and $0$ otherwise.
During RL post-training, the goal is to learn a policy maximizing the expected discounted return (\ie the $Q$-function), which we re-write with respect to the observation-conditioned policy for simplicity,
$Q^\pi(\vo,\va)=\E_\pi\left[ \sum_{t=0}^{\infty} \gamma^t r_t | \vo_0=\vo, \va_0=\va \right]$.
In robot manipulation, this objective corresponds to solving tasks with both high success rate and high efficiency.

\subsection{Flow-Matching Base Policies From Imitation}

Although the proposed adaptation framework is not restricted to a particular policy parameterization, we instantiate it with flow-matching (FM) policies~\cite{liu2023flow} throughout this work. We choose FM models because they retain the expressive action generation capabilities of diffusion-based policies~\cite{ho2020ddpm,song2021ddim,chi2023diffusion}, while being simpler to implement and requiring only a small number of iterative steps at inference time, enabling high-frequency policy execution.
We assume that the base policy consists of an observation encoder $\gE$ and a conditional flow model. Given an observation $\vo$, the encoder produces a latent conditioning vector $\vc = \gE(\vo)$, which is then used by the flow model to generate actions. This factorization is common in modern policies~\cite{lbm2025careful,chi2023diffusion,lei2025rl100}, where perception first extracts a representation and action generation is performed conditioned on that representation.

Given an offline dataset $\gD_{\mathrm{off}}=\{\tau_i\}_{i=1}^N$, where each trajectory $\tau_i=\{(\vo_t,\va_t)\}_{t=1}^{H_i}$ contains observation--action pairs, the FM policy is trained by learning a velocity field $v_\phi$ for an ordinary differential equation (ODE) that transports a noise sample toward the target action.
Specifically, let $\va^0=\va$ denote the clean action and $\va^1=\vw \sim \gN(\bm{0},\mI)$ denote a Gaussian noise sample. For an interpolation variable $k\sim \mathrm{Unif}[0,1]$, we define the intermediate point $\va^k = (1-k)\va^0 + k\va^1$.
The imitation learning objective is then
\begin{equation}
\label{eq:il}
\gL_{\mathrm{IL}}(\phi)
=
\E_{\substack{(\vo,\va)\sim\gD_{\mathrm{off}}\\
k\sim\mathrm{Unif}[0,1]\\
\vw\sim\gN(\bm{0},\mI)}}
\left[
\left\|
v_\phi(\va^k,k,\vc) - (\vw-\va)
\right\|_2^2
\right],
\end{equation}
where $\vc=\gE_\phi(\vo)$. The encoder $\gE_\phi$ and the velocity model $v_\phi$ are optimized jointly during imitation learning.

During deployment, actions are generated by numerically integrating the learned flow from noise to action using a reversed discrete schedule $1=k_M > \dots > k_m > \dots > k_0=0$.
Starting from $\va^{1}=\vw \sim \gN(\bm{0},\mI)$, the action is obtained by
\begin{equation}
\label{eq:fm_sample}
\begin{aligned}
\va^{k_m} &= \va^{k_{m+1}} - (k_{m+1}-k_m)v_\phi(\va^{k_{m+1}},k_{m+1},\vc),\\
m&=M-1,\dots,0.
\end{aligned}
\end{equation}
The final sample $\va^0$ is taken as the generated action $\va_\mathrm{off}$. We denote the full policy by $\pi_{\mathrm{base}}(\cdot | \gE(\vo))$.
Note that, throughout this paper, actions refer to a multi-step chunk denoted by $\va$ and we denote a single-step action by $a$.



\section{Steering Policies with Z-Perturbation Reinforcement Learning}

\begin{figure*}[!t]
    \centering
    \includegraphics[width=\textwidth]{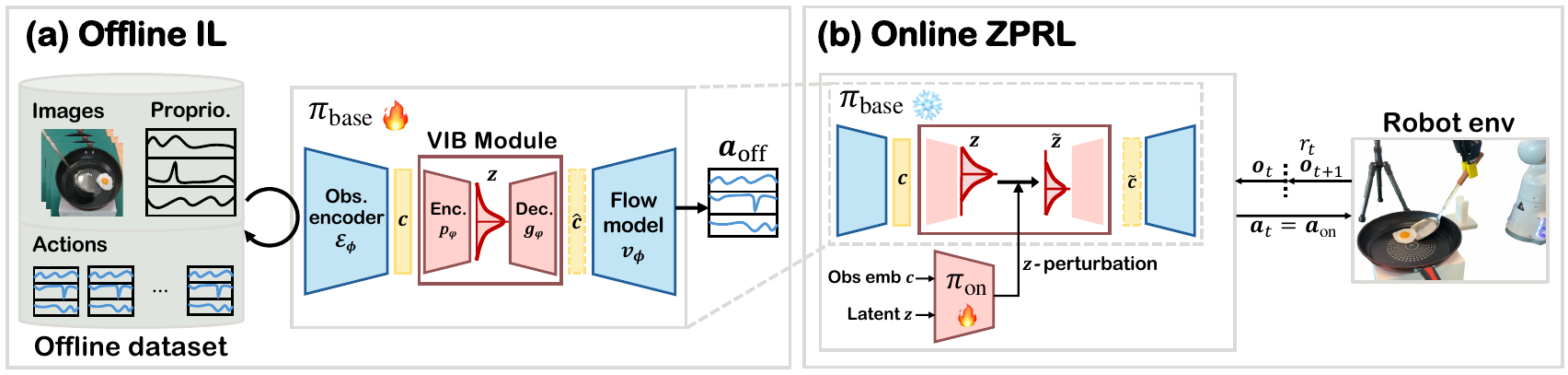}
    \par\vspace*{-3mm}
    \caption{\textbf{Two-stage training pipeline of ZPRL.} (a) Offline, a flow-based manipulation policy is pretrained with a VIB bottleneck over the task-conditioning embedding. (b) Online, the pretrained backbone is frozen and a latent residual policy predicts $\Delta\vz$ to perturb the bottleneck, $\tilde{\vz}=\vz+\lambda\Delta\vz$, thereby steering the generated action through the frozen VIB decoder and flow policy.}
    \par\vspace*{-5mm}
    \label{fig:zprl_pipeline}
\end{figure*}

We present Z-Perturbation Reinforcement Learning (ZPRL), a lightweight post-training framework for adapting pretrained manipulation policies through online RL. Our key idea is to intervene on an internal task-relevant interface of the base policy, rather than on policy weights or output actions.
Concretely, we first augment the base policy with a VIB module at the boundary where the observation representation conditions the action expert. Thus we can obtain a compact bottleneck latent from the observation embedding, following prior work on structured representations for iterative imitation~\cite{jin2025soe}.
We then learn an online residual policy that perturbs this latent during post-training, while keeping the entire base policy frozen. The perturbed latent is then passed through the VIB decoder and used as the condition of the flow model, realizing ``policy steering'' through a modified condition.
\Figref{fig:zprl_pipeline} illustrates the overall two-stage pipeline, including offline bottlenecked IL and online residual perturbation.

\subsection{Bottleneck Task Latent for Policy Conditioning}

A central challenge in post-training manipulation policies is to find an efficient interface for behavior steering. Direct exploration in action space becomes increasingly difficult as the action horizon grows, since the dimensionality of the action chunk scales linearly with time. In such spaces, adding Gaussian noise inevitably brings high-frequency jitter rather than meaningful exploratory diversity. A more suitable interface should therefore preserve action-relevant structure while discarding unnecessary details in the observation or its embedding. Following Deep VIB~\cite{alemi2017deep} and SOE~\cite{jin2025soe}, we adopt a compact bottleneck latent on top of the observation embedding. The VIB objective encourages the suppression of task-irrelevant variation while preserving what the action expert needs. It is more of a regularizer than an invariance guarantee. We formulate the information bottleneck objective:
\begin{equation}
\label{eq:ib_obj}
\min \; - I(\rvz;\rva) + \beta I(\rvz;\rvc),
\end{equation}
where $I$ denotes mutual information between random variables, $\vc=\gE(\vo)$ is the observation embedding produced by the base encoder, and $\beta$ controls the trade-off between informativeness and compactness.

To optimize this objective in practice, we instantiate it with a variational encoder-decoder parameterization following prior work~\cite{alemi2017deep, jin2025soe}. This yields the standard VIB surrogate objective
\begin{equation}
\label{eq:vib_ub_clean}
\begin{aligned}
\gL_{\mathrm{VIB}}(\varphi) = \E_{(\vo,\va)\sim\gD}
\Big[
&-\E_{\vz\sim p_\varphi(\cdot | \vc)}
\log q(\va | g_\varphi(\vz)) \\
&+ \beta\,\mathrm{KL}\!\left(p_\varphi(\rvz | \vc) \| r(\rvz)\right)
\Big],
\end{aligned}
\end{equation}
where the VIB encoder $p_\varphi(\cdot | \vc)$ maps the observation embedding to a latent posterior, and the VIB decoder $g_\varphi(\vz)$ reconstructs a feature for downstream action generation. For computational tractability, we parameterize the posterior as a diagonal Gaussian with mean and variance predicted by the VIB encoder. The latent prior is chosen as a standard Gaussian $r(\rvz)=\mathcal N(\bm{0},\mI)$.

In our flow-policy instantiation, the first term in~\Eqref{eq:vib_ub_clean} is realized by the same flow-matching objective used for behavior cloning, with the reconstructed feature $\hat{\vc}=g_\varphi(\vz)$ replacing the original observation embedding $\vc$ as the conditioning input. Therefore, the VIB loss becomes
\begin{equation}
\label{eq:vib_flow}
\begin{aligned}
\gL_{\mathrm{VIB}}(\varphi) =
\E_{\substack{(\vo,\va),\\
k,\vw,\vz}}
\Big[
&\left\| v_\phi(\va^k,k,\hat{\vc})-(\vw-\va)\right\|_2^2 \\
&+\beta\,\mathrm{KL}\!\left(p_\varphi(\rvz | \vc)\,\|\,r(\rvz)\right)
\Big],
\end{aligned}
\end{equation}
where $\vc=\gE_\phi(\vo)$ and $\hat{\vc}=g_\varphi(\vz)$. For joint training (\ie updating the base and VIB concurrently), the offline loss is
\begin{equation}
\label{eq:offline_obj}
\gL_{\mathrm{off}}(\phi,\varphi)=\gL_{\mathrm{IL}}(\phi)+\gL_{\mathrm{VIB}}(\varphi).
\end{equation}
As in~\cite{jin2025soe}, we treat the bottleneck as a plug-in auxiliary path: gradients from $\gL_{\mathrm{VIB}}$ are blocked from updating the base encoder $\gE_\phi$ and flow generator $v_\phi$, so the original imitation path remains unaffected. This preserves the performance of the base policy while learning a compact latent interface.
Alternatively, the VIB can be fitted post hoc by freezing an independently trained base policy and optimizing only $\varphi$ with~\Eqref{eq:vib_flow}. This plug-and-play operation requires accessible intermediate features and task-relevant offline observation-action pairs. The original base-policy training dataset is not required (Appendix~\ref{app:posthoc}).

ZPRL inherits the VIB interface, auxiliary dual path, and stop-gradient design from SOE~\cite{jin2025soe}. SOE steers this latent by changing posterior variance and manually selecting dimension indices, then uses collected rollouts to retrain the policy. ZPRL instead makes steering reward-driven: an actor learns state-conditioned residual directions and magnitudes, a critic evaluates the perturbed latent under task reward, and the entire base policy remains frozen.

\subsection{RL Perturbation on Bottleneck Latents}

Rather than overwriting the bottleneck latent learned offline, we adapt the pretrained policy by applying a residual perturbation to it. Specifically, given the current bottleneck representation $\vz$, the online policy predicts a latent correction $\Delta \vz$ and forms a perturbed latent
\begin{equation}
   \label{eq:perturb}
   \tilde{\vz} = \vz + \lambda \Delta \vz,
   \qquad
   \Delta \vz \sim \pi_{\mathrm{on}}(\cdot | \vc, \vz),
\end{equation}
where $\pi_{\mathrm{on}}$ is the online RL policy and $\lambda$ controls the perturbation magnitude. The perturbed latent is then decoded into a new conditioning embedding $\tilde{\vc} = g_{\varphi}(\tilde{\vz})$, which replaces the original condition used by the flow policy. The final action is therefore generated as $\va_{\mathrm{on}} \sim \pi_{\mathrm{base}}(\cdot | \tilde{\vc}),$
so that online RL steers the base policy by modifying its internal task-conditioning variable before action generation. The perturbed control loop is illustrated in \Figref{fig:zprl_pipeline}(b).

ZPRL performs within-support online policy steering. It assumes that offline training has supplied the behavioral primitives required for success, and remaining failures mainly arise from behavior selection, timing, and composition. This does not require a base policy with high success rate: necessary motions may be represented but invoked in the wrong states or at the wrong time. Online RL uses reward to improve their deployment through latent perturbations, while the VIB encourages task-relevant compression and the frozen decoder and flow generator preserve the pretrained action prior.

During online learning, rewards and values should be associated with the \emph{resulting perturbed latent} rather than the residual $\Delta \vz$ alone. We therefore define
\begin{equation}
\label{eq:reward_q}
\hat r(\vc,\tilde{\vz}) \coloneqq r(\vo, \va_{\mathrm{on}}),
\qquad
\hat Q(\vc,\tilde{\vz}) \coloneqq Q(\vo, \va_{\mathrm{on}}),
\end{equation}
where $\va_{\mathrm{on}} \sim \pi_{\mathrm{base}}(\cdot | g_{\varphi}(\tilde{\vz})).$ The critic is defined on $\tilde{\vz}$ rather than on $\Delta \vz$ because the same residual can induce different action distributions when added to different $\vz$, due to the stochasticity of $p_{\varphi}(\vz | \vc)$. Therefore, $\Delta \vz$ alone is insufficient to determine the downstream action and cannot be assigned a well-defined value without its latent base. In our implementation, the actor takes $(\vc,\vz)$ as input to predict $\Delta\vz$, while the critic evaluates the resulting pair $(\vc,\tilde{\vz})$.

The bottleneck latent provides a compact, task-aligned control interface. First, it influences the whole action chunk through a latent of fixed dimensions (mostly 8, 16, 32 in this work), independent of the chunk length. Second, the perturbation acts on the policy condition rather than directly on the generated action, so the final action is still produced by the pretrained generator and remains structured by the offline data distribution~\cite{he2026demystifying}. In this way, online RL steers the action generation process instead of bypassing it, which can lead to safer and more stable actions in the environment.

\subsection{Online RL Objective and Design Choices}

In principle, ZPRL can use various RL algorithms; we instantiate it with soft actor-critic (SAC)~\cite{haarnoja2019soft}. To simplify notation, we drop the subscript $\mathrm{on}$ and denote the online perturbation policy by $\pi_\theta$. During online adaptation, only the residual actor $\pi_\theta$ and the latent critic $Q_\psi$ are updated, while the pretrained base policy remains frozen.

Given a latent state $(\vc,\vz)$, the actor predicts a stochastic residual $\Delta \vz \sim \pi_\theta(\cdot \mid \vc, \vz)$, which forms the perturbed latent $\tilde{\vz} = \vz + \lambda \Delta \vz$. The actor is optimized to maximize returns while maintaining sufficient exploration:
\begin{equation}
   \label{eq:policy_loss}
   \gL_{\pi}(\theta)
   =
   \E_{\substack{(\vc,\vz)\sim\gD_{\mathrm{on}}\\
   \Delta\vz\sim\pi_\theta(\cdot\mid \vc,\vz)}}
   \Big[
   \alpha \log \pi_\theta(\Delta\vz\mid \vc,\vz)
   -
   Q_\psi(\vc,\tilde{\vz})
   \Big],
\end{equation}
where $\alpha$ is the temperature coefficient and is adjusted automatically. The critic is trained with a temporal-difference objective,
\begin{equation}
   \label{eq:q_loss}
   \gL_{Q}(\psi)
   =
   \E_{\substack{(\vc,\vz,r,\vc',\vz')\sim\gD_{\mathrm{on}}\\
   \Delta\vz'\sim\pi_\theta(\cdot\mid \vc',\vz')}}
   \Big[
   \big(
   Q_\psi(\vc,\tilde{\vz}) - y
   \big)^2
   \Big],
\end{equation}
with target
\begin{equation}
   y = r + \gamma \,\bar{Q}_{\bar{\psi}}(\vc',\tilde{\vz}'),
   \qquad
   \tilde{\vz}' = \vz' + \lambda \Delta\vz',
\end{equation}
where $\bar{Q}_{\bar{\psi}}$ denotes the target critic updated by exponential moving average of $Q_\psi$. Additional implementation details are provided in Appendix~\ref{app:details}. Here we use a variant without the entropy term in the Q function to stabilize training.

\noindent\textbf{Choosing the perturbation scale $\lambda$.}
The perturbation scale $\lambda$ controls the strength of online steering in the bottleneck latent space. A small $\lambda$ has little effect on the latent and adaptation remains weak; if it is too large, the perturbed latent may move too far from the pretrained latent distribution and become harder for the VIB decoder and action prior to handle reliably. In practice, we track the magnitude of the original latent $\vz$ and the perturbation $\Delta \vz$ to determine a fixed $\lambda$ for each task. We study its effects in \Secref{sec:exp_z} and provide practical guidelines and task-specific values in Appendix~\ref{app:lambda}.

\noindent\textbf{Update-to-data ratio.}
A high update-to-data (UTD) ratio (the ratio between number of gradient steps and robot transitions) can improve performance measured against environment steps, but it does not necessarily improve wall-clock efficiency. In our setting, increasing the number of gradient updates per action chunk quickly shifts the computational bottleneck from robot interaction to policy updates, especially when multiple $Q$ functions are used for stability~\cite{chen2021redq}. Since our goal is efficient online adaptation in real time rather than only step efficiency, we do not use an excessively high UTD ratio (such as 10 or 20) in experiments. Instead, we use $\mathrm{UTD}=1$ in simulations and 2 or 5 in real-world tasks to balance sample efficiency and wall-clock throughput.


\begin{figure*}[!t]
    \centering
    \includegraphics[width=0.24\textwidth]{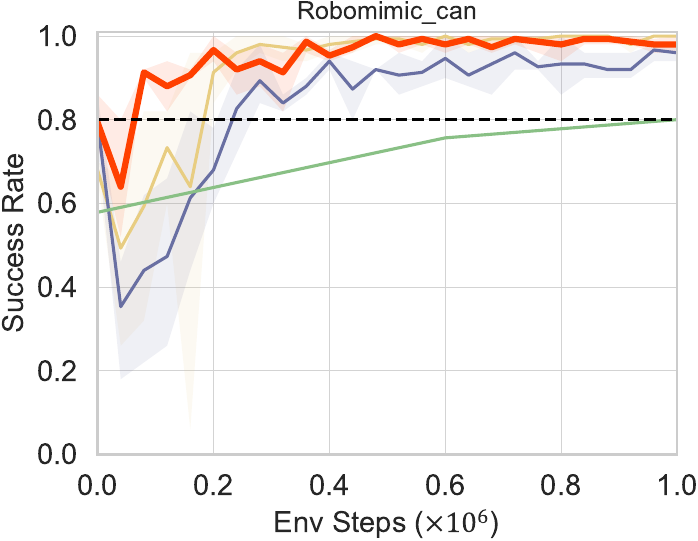}\hfill
    \includegraphics[width=0.24\textwidth]{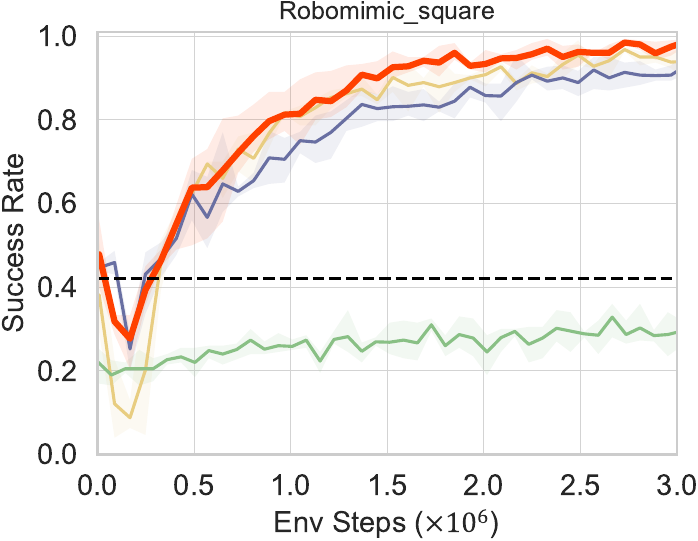}\hfill
    \includegraphics[width=0.24\textwidth]{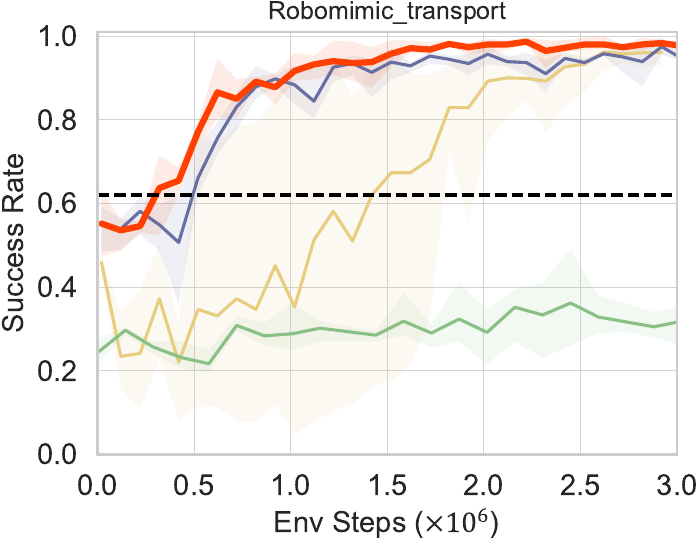}\hfill
    \includegraphics[width=0.24\textwidth]{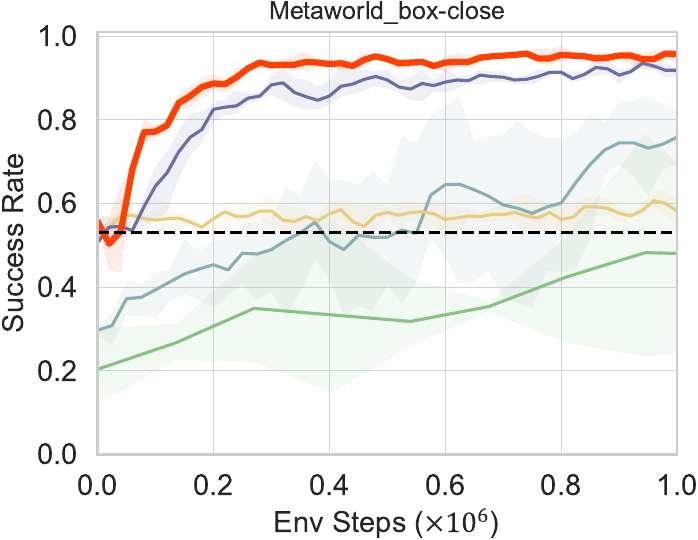}\\[1mm]
    \includegraphics[width=0.24\textwidth]{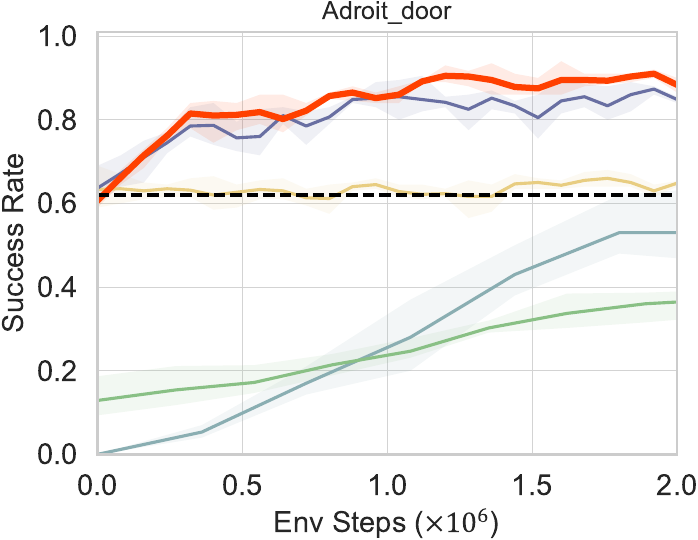}\hfill
    \includegraphics[width=0.24\textwidth]{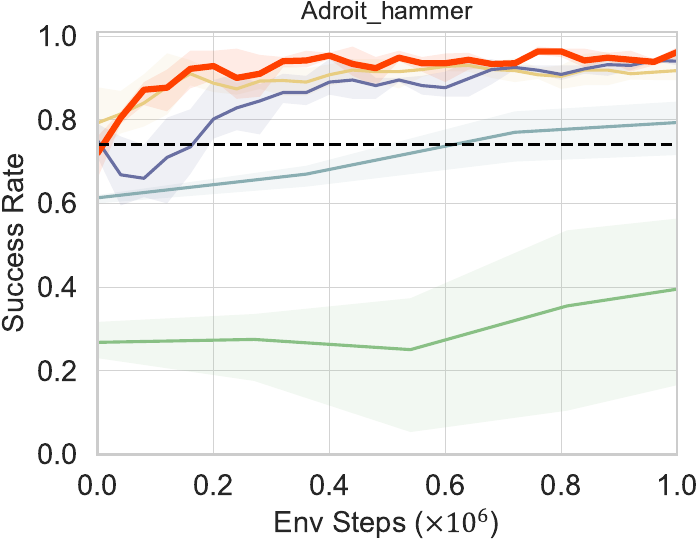}\hfill
    \includegraphics[width=0.24\textwidth]{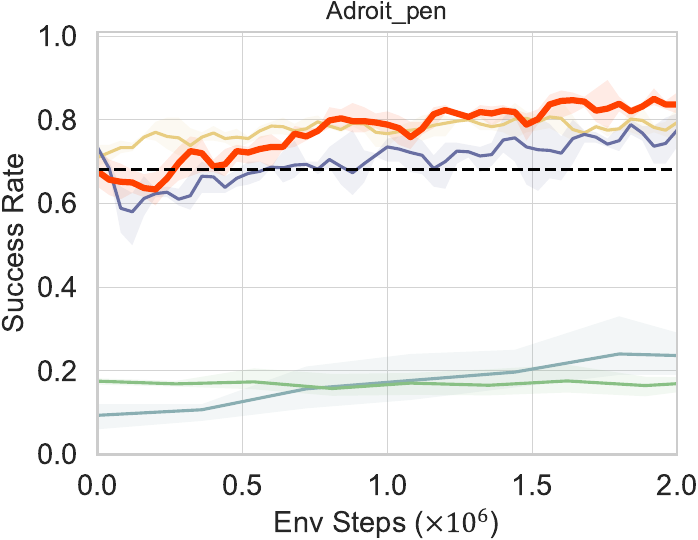}\hfill
    \includegraphics[width=0.24\textwidth]{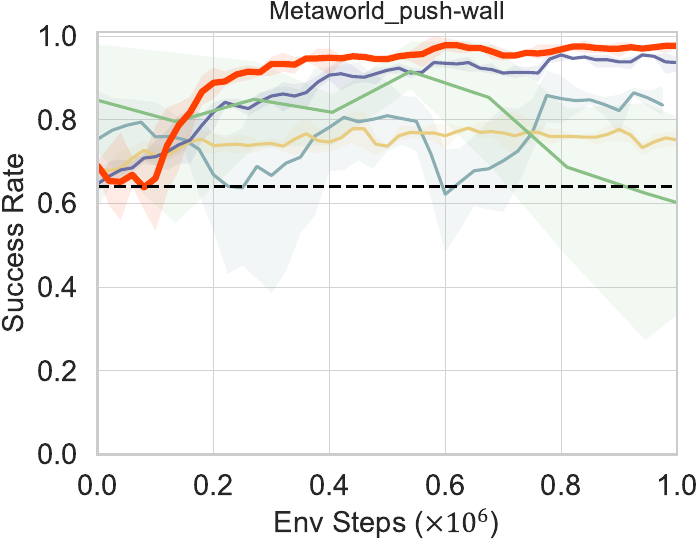}
    \vspace{2pt}
    \includegraphics[width=0.55\textwidth]{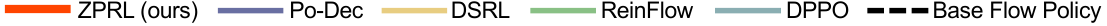}
    \par\vspace*{-3mm}
    \caption{\textbf{Simulation results across three benchmarks.} Success rate versus online environment steps during finetuning. Unless otherwise specified, all online curves in simulation are averaged over 3 random seeds, and shaded regions indicate the 95\% interval across seeds. Evaluation protocols per task are given in Appendix~\ref{app:arch}.}
    \vspace{-3mm}
    \label{fig:sim_main}
\end{figure*}

\section{Experiments}

We evaluate ZPRL on eight tasks across three simulation benchmarks and compare it against several RL finetuning baselines in \Secref{sec:sim}. We then investigate how the bottleneck latent inherited from the pretrained base policy influences online finetuning in \Secref{sec:exp_z}. Next, we compare the success-smoothness trade-off of ZPRL with a residual RL method and its smoothed variants in \Secref{sec:exp_smooth}. Finally, we demonstrate real-world ZPRL on four robotic tasks in \Secref{sec:real} and discuss the limitations and possible extensions of the proposed method in \Secref{sec:limitation}.

\subsection{Online RL Finetuning in Simulation}
\label{sec:sim}

We first compare ZPRL against recent methods for RL adaptation of IL base policies. We consider eight tasks from three benchmarks: \texttt{Can}, \texttt{Square}, and \texttt{Transport} from \texttt{Robomimic}~\cite{Mandlekar2022robomimic}; \texttt{Door}, \texttt{Hammer}, and \texttt{Pen} from \texttt{Adroit}~\cite{Kumar2016adroit}; and \texttt{Box-close} and \texttt{Push-wall} from \texttt{Metaworld}~\cite{mclean2025metaworld}. Task selection balances room for improvement, skill coverage, and availability of datasets and baseline results (Appendix~\ref{app:selection}). We focus on vision-based manipulation, where observations include both images and proprioception. For each task, all base policies are trained on the same offline dataset: for \texttt{Robomimic}, 100 trajectories from the official mixed-quality multi-human (MH) dataset released by~\cite{Mandlekar2022robomimic}; for \texttt{Adroit} and \texttt{Metaworld}, 100 trajectories generated by a medium-expert policy from~\cite{xu2024drm}. We compare with four baselines: DPPO~\cite{ren2025dppo} and ReinFlow~\cite{zhang2025reinflow}, which finetune the full diffusion or flow policies with on-policy optimization; Policy-Decorator (Po-Dec)~\cite{yuan2025policy}, a representative action-residual method with progressive exploration and scaled residuals for stable learning; and DSRL~\cite{wagenmaker2025steering}, implemented as DSRL-SAC to steer the base policy through its input noise $\vw$. Baseline configurations and log sources are given in Appendix~\ref{app:baselines}.

As shown in~\Figref{fig:sim_main}, ZPRL consistently reaches strong final performance across all eight tasks and is among the most effective methods in terms of online adaptation speed. This holds for both parallel-jaw manipulation (\texttt{Robomimic}, \texttt{Metaworld}) and dexterous hands (\texttt{Adroit}), suggesting that bottleneck latent perturbation provides a broadly effective interface for improving pretrained policies with RL. In particular, ZPRL achieves high asymptotic returns while remaining competitive in sample efficiency on most tasks.
Although DSRL reaches asymptotic SRs comparable to other effective methods on \texttt{Robomimic}, its gains on \texttt{Metaworld} remain modest under our tuned configuration.

On \texttt{Metaworld}, we use $\dim(\vz)=8$, matching the dimension of the two-step action chunk (four dimensions per step). This task-specific latent capacity is supported by the \texttt{Push-wall} sweep in Appendix~\ref{app:cross_task}. ZPRL reaches high SR earlier than Po-Dec on both tasks, highlighting that efficiency benefit comes not only from a small action space but also the intervention interface.

Po-Dec is a competitive baseline with comparable final SRs across the eight tasks, while ZPRL generally reaches high SR with fewer interactions (see statistics in Appendix~\ref{app:interaction}). For example, on \texttt{Box-close}, ZPRL and Po-Dec first reach SR=0.95 at 0.20M and 0.53M steps, respectively. Their success-smoothness trade-off is examined in~\Secref{sec:exp_smooth}.

\subsection{How the Bottleneck Latent Interface Shapes Online RL}
\label{sec:exp_z}

\begin{figure*}[!t]
    \centering
    \vspace{0pt}
    \includegraphics[width=0.24\textwidth]{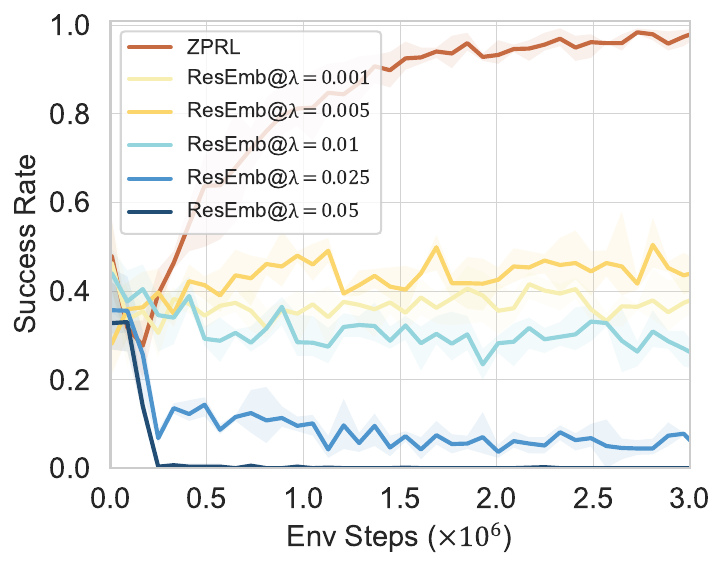}
    \includegraphics[width=0.24\textwidth]{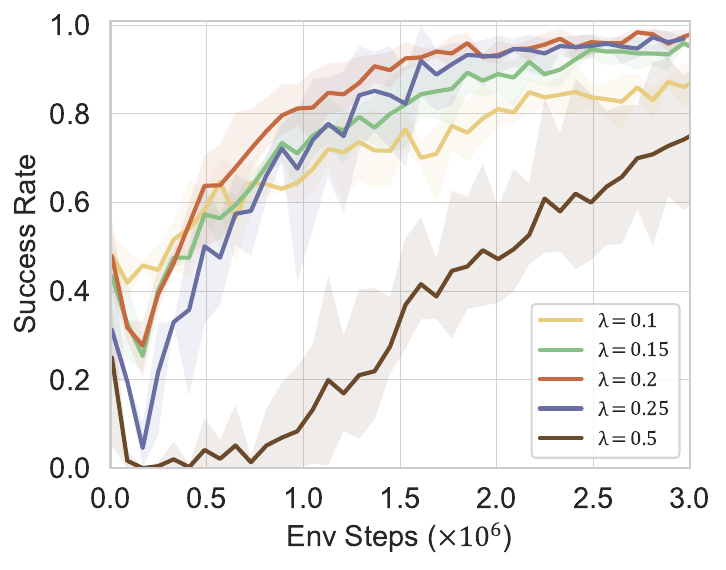}\hfill
    \includegraphics[width=0.24\textwidth]{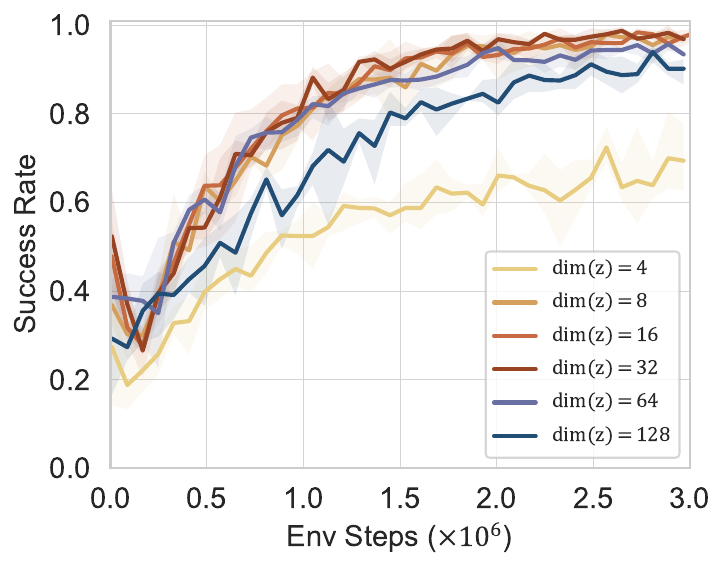}\hfill
    \includegraphics[width=0.24\textwidth]{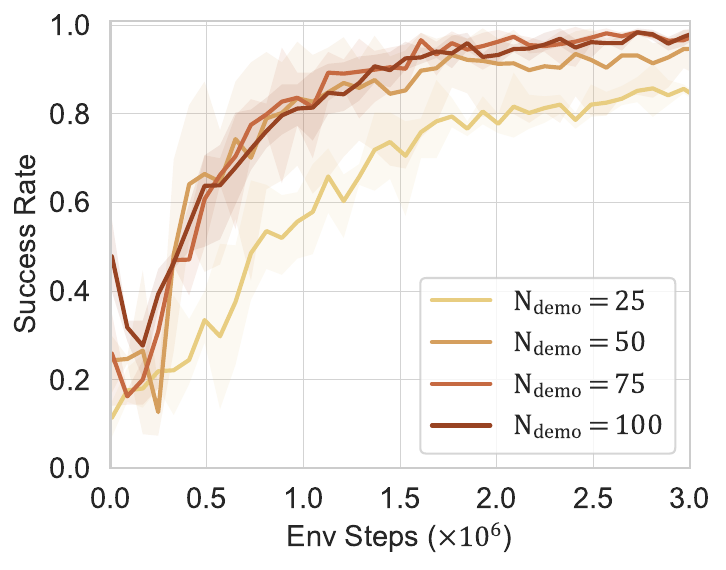}
    \par\vspace*{-3mm}
    \caption{\textbf{ZPRL with different latent interface settings on \texttt{Square} in \texttt{Robomimic}.} We study (from left to right): (a) direct perturbation on the observation embedding (\textsc{ResEmb}); (b) the perturbation scale $\lambda$; (c) the dimension of $\vz$; and (d) the number of trajectories in the offline dataset.}
    \vspace*{-2mm}
    \label{fig:sim_ablation}
\end{figure*}

To understand whether and how the bottleneck latent serves as an effective interface for online finetuning, we perform four ablations on the \texttt{Square} task in \texttt{Robomimic}. Its base SR of about 40\% leaves substantial room for online improvement, and using the same task isolates design choices. We study: (a) the type of steering interface, where we compare ZPRL against directly learning residuals on the observation embedding without VIB compression; (b) the perturbation scale $\lambda$, which controls how far RL can move around the pretrained latent; (c) the bottleneck dimension $\dim(\vz)$, which determines how compact the steering space is; and (d) the number of offline demonstrations used to train the base policy, which affects how well the latent manifold is supported by data. Cross-task ablations in Appendix~\ref{app:cross_task} examine whether these trends extend beyond \texttt{Square}.

\begin{table}[!ht]
\caption{Summary metrics of online finetuning under different latent ablations on \texttt{Square}. Final success rates (SRs) are averaged over the last 5 data points. N/A means no seeded run of the variant reaches the threshold.}
\label{tab:ablation_summary}
\par\vspace*{2mm}
\centering
\begin{tabular}{lcc}
\toprule
Setting         & Steps ($\times10^6$) to SR=0.9 $\downarrow$ & Final SR $\uparrow$ \\
\midrule
$\lambda=0.10$  & 2.40 & 0.86 \\
$\lambda=0.15$  & 1.54 & 0.95 \\
$\lambda=0.20$  & 1.16 & 0.98 \\
$\lambda=0.25$  & 1.15 & 0.96 \\
$\lambda=0.50$  & N/A  & 0.74 \\
\midrule
$\dim(\vz)=4$   & N/A  & 0.68 \\
$\dim(\vz)=8$   & 1.23 & 0.98 \\
$\dim(\vz)=16$  & 1.16 & 0.98 \\
$\dim(\vz)=32$  & 1.08 & 0.99 \\
Po-Dec ($\dim(\va)=40$)  & 1.68 & 0.92 \\
$\dim(\vz)=64$  & 1.29 & 0.94 \\
$\dim(\vz)=128$ & 1.90 & 0.91 \\
\midrule
$N_{\mathrm{demo}}=25$   & 2.54 & 0.86 \\
$N_{\mathrm{demo}}=50$   & 1.21 & 0.96 \\
$N_{\mathrm{demo}}=75$   & 1.06 & 0.98 \\
$N_{\mathrm{demo}}=100$  & 1.16 & 0.98 \\
\bottomrule
\end{tabular}
\vspace{-3mm}
\end{table}

\noindent\textbf{Bottleneck latent versus direct steering on observation embeddings.}
We first compare ZPRL with \textsc{ResEmb}, which directly learns residuals on the high-dimensional observation embedding without VIB compression. \textsc{ResEmb} performs consistently worse and becomes more unstable as residual scale grows, suggesting that effective online adaptation does not arise from steering arbitrary intermediate features. A likely reason is that the raw observation embedding still contains task-irrelevant information, which makes RL exploration harder and less focused. The VIB encourages task-relevant compression, providing a compact space in which RL can steer action generation more efficiently.

A deterministic bottleneck also supports successful online learning (Appendix~\ref{app:deterministic}; KL-weight sensitivity sweep in Appendix~\ref{app:beta}), indicating that variational regularization is not always necessary. It is the compact, task-aligned bottleneck that actually matters. We take the stochastic version as the default to indicate inheritance from SOE~\cite{jin2025soe}.

\noindent\textbf{Perturbation scale $\lambda$ controls the exploration-stability trade-off.}
As shown in~\Figref{fig:sim_ablation}(b), the perturbation scale $\lambda$ affects both the speed of online improvement and the final success rate (SR). A larger $\lambda$ gives the RL policy stronger ability to steer the pretrained policy, but it also makes the randomly initialized $\Delta \vz$ more likely to push $\tilde{\vz}$ away from the local neighborhood of the pretrained latent. This weakens the support provided by the frozen action prior and can substantially destabilize finetuning. In the extreme case of $\lambda=0.5$, the policy only achieves a final SR of 0.74. 
In contrast, a smaller $\lambda$ keeps the perturbation more local and leads to more stable optimization, but also restricts how quickly and how strongly RL can modify behavior. This is reflected in the slower learning curves of $\lambda=0.10$ and $0.15$. Empirically, $\lambda=0.20$ provides the best balance between adaptability and stability on \texttt{Square}.
These results support our central design principle that online RL should steer the pretrained latent locally through residual perturbations, rather than overwrite the latent interface with overly large corrections.

\noindent\textbf{The gain is not merely from dimensionality reduction.}
We vary the bottleneck dimension from 4 to 128, while adjusting $\lambda$ according to the magnitude of $\vz$. As shown in~\Figref{fig:sim_ablation}(c) and Table~\ref{tab:ablation_summary}, ZPRL is robust to a moderate range of bottleneck sizes on \texttt{Square}: $\dim(\vz)=8,16,32$ all achieve strong online performance with similar steps to reach SR=0.9 and similar final SRs. The best result is obtained at $\dim(\vz)=32$.
Importantly, the gain of ZPRL cannot be explained simply by using a lower-dimensional control variable than the action chunk. Even when $\dim(\vz)=64$, which already exceeds the action dimension of Po-Dec ($\dim(\va)=40$), ZPRL still reaches SR$=0.9$ faster and achieves a higher final SR. This suggests that what matters is not dimension reduction alone, but whether RL operates in a task-relevant space. As long as perturbations remain local, the frozen VIB decoder and action generator can still map corrected latents to meaningful actions.
At the same time, the two extremes both hurt performance. When $\dim(\vz)=4$, the bottleneck is too restrictive and likely discards useful task information, leading to poor online learning. When $\dim(\vz)=128$, the steering space becomes less compact and harder to optimize efficiently, reducing both learning speed and final performance. Overall, these results suggest that a useful steering interface should be compact, but not so restrictive that it removes information needed for success. Such phenomenon can also be observed in~\cite{pertsch2020accelerating}.

\noindent\textbf{The effect of offline dataset size is task-dependent.}
On \texttt{Square}, reducing the offline dataset to 25 demonstrations slows online learning and lowers final SR, although ZPRL remains effective (\Figref{fig:sim_ablation} and Table~\ref{tab:ablation_summary}). In contrast, the dataset-size variants on \texttt{Adroit\_door} perform similarly (Appendix~\ref{app:cross_task}). This may reflect the more homogeneous data generated by a single policy checkpoint on \texttt{Door}, compared with the mixed-quality multi-human data in \texttt{Robomimic}. Thus, data coverage and diversity, rather than trajectory count alone, determine the benefit of additional offline data.

\textbf{Takeaway.}
These ablations suggest that the bottleneck latent shapes ZPRL through several properties: locality of intervention, compactness of the control space, coverage of the pretrained latent manifold, and the form of the steering interface itself. In particular, the comparison with direct residual steering on the observation embedding shows that the gain does not come from perturbing an arbitrary intermediate feature. These results support a compact, task-aligned bottleneck as an effective interface for local steering within pretrained behavioral support.

\subsection{Smooth and Structured Action Generation}
\label{sec:exp_smooth}


We evaluate the base policy, Po-Dec, its training- or test-time smoothed variants, and ZPRL on the same 150 initial layouts of \texttt{Square}, with all policy stochasticity disabled. SR is computed over all layouts. Smoothness is measured from observed end-effector positions on successful trajectories, accounting for the low-level controller's response to commanded actions. We use finite differences at the control interval $\Delta t=0.05$\,s. For position $\mathbf{p}_t$, the translational acceleration and jerk norms are
\begin{equation}
\begin{aligned}
\mathrm{Acc}_t &= \left\|\frac{\mathbf{p}_{t+2}-2\mathbf{p}_{t+1}+\mathbf{p}_t}{\Delta t^2}\right\|_2,\\
\mathrm{Jerk}_t &= \left\|\frac{\mathbf{p}_{t+3}-3\mathbf{p}_{t+2}+3\mathbf{p}_{t+1}-\mathbf{p}_t}{\Delta t^3}\right\|_2.
\end{aligned}
\end{equation}
Differences are computed between control-step samples in successful trajectories. Then the mean and sample standard deviation are obtained from these samples. Each method contributes more than 10,000 control-step samples per metric.

Po-Dec's test-time filter uses a four-step window to get an exponentially weighted moving average (EMA), $\hat a_t=(\sum_{i=0}^{3}\omega^i a_{t-i})/\sum_{i=0}^{3}\omega^i$, with $\omega\in\{0.6,0.7,0.8,0.9\}$. The training-time CAPS (Conditioning for Action Policy Smoothness) regularizer~\cite{mysore2021caps} uses $(\lambda_s,\lambda_t,\sigma)=(0.1,0.1,0.01)$ for the low setting and $(5.0,1.0,0.02)$ for the high setting to consider spatial and temporal smoothness during policy learning. ZPRL uses neither temporal filtering nor smoothness regularization.

\begin{table}[!t]
    \caption{\textbf{Comparison of robot state smoothness across policy variants.} SR uses all 150 layouts; acceleration and jerk norms are reported as mean~$\pm$ standard deviation over control steps of successful trajectories.}
    \label{tab:smooth_metric}
    \centering
    \footnotesize
    \setlength{\tabcolsep}{2pt}
    \renewcommand{\arraystretch}{1.05}
    \begin{tabular}{@{}p{1.50in} c c c@{}}
        \toprule
        \makecell[l]{Method} &
        \makecell[c]{SR} &
        \makecell[c]{Translational\\acc. $(\si{m/s^2})$} &
        \makecell[c]{Translational\\jerk $(\si{m/s^3})$} \\
        \midrule

        \textbf{Base}
        & 37.3\%
        & $0.4 \pm 0.3$
        & $10.0 \pm 6.8$ \\

        \midrule

        \textbf{Po-Dec}
        & 90.7\%
        & $1.2 \pm 0.7$
        & $32.3 \pm 19.6$ \\

        Po-Dec + EMA ($\omega=0.6$)
        & 88.7\%
        & $1.0 \pm 0.7$
        & $28.4 \pm 19.3$ \\

        Po-Dec + EMA ($\omega=0.7$)
        & 90.0\%
        & $1.0 \pm 0.7$
        & $28.0 \pm 19.4$ \\

        Po-Dec + EMA ($\omega=0.8$)
        & 90.0\%
        & $1.0 \pm 0.7$
        & $27.5 \pm 19.3$ \\

        Po-Dec + EMA ($\omega=0.9$)
        & 89.3\%
        & $1.0 \pm 0.7$
        & $27.5 \pm 19.4$ \\

        \midrule

        \makecell[l]{\textbf{CAPS} low\\[-1pt]
        \scriptsize$(\lambda_s=0.1,\lambda_t=0.1,\sigma=0.01)$}
        & 96.0\%
        & $1.1 \pm 0.6$
        & $29.7 \pm 18.3$ \\

        \makecell[l]{\textbf{CAPS} high\\[-1pt]
        \scriptsize$(\lambda_s=5.0,\lambda_t=1.0,\sigma=0.02)$}
        & 73.3\%
        & $0.8 \pm 0.5$
        & $19.2 \pm 13.2$ \\

        \midrule

        \textbf{ZPRL}
        & 98.7\%
        & $0.9 \pm 0.6$
        & $19.7 \pm 14.5$ \\

        \bottomrule
    \end{tabular}
    \vspace{-2mm}
\end{table}

\begin{figure}[!ht]
    \centering
    \includegraphics[width=0.95\columnwidth]{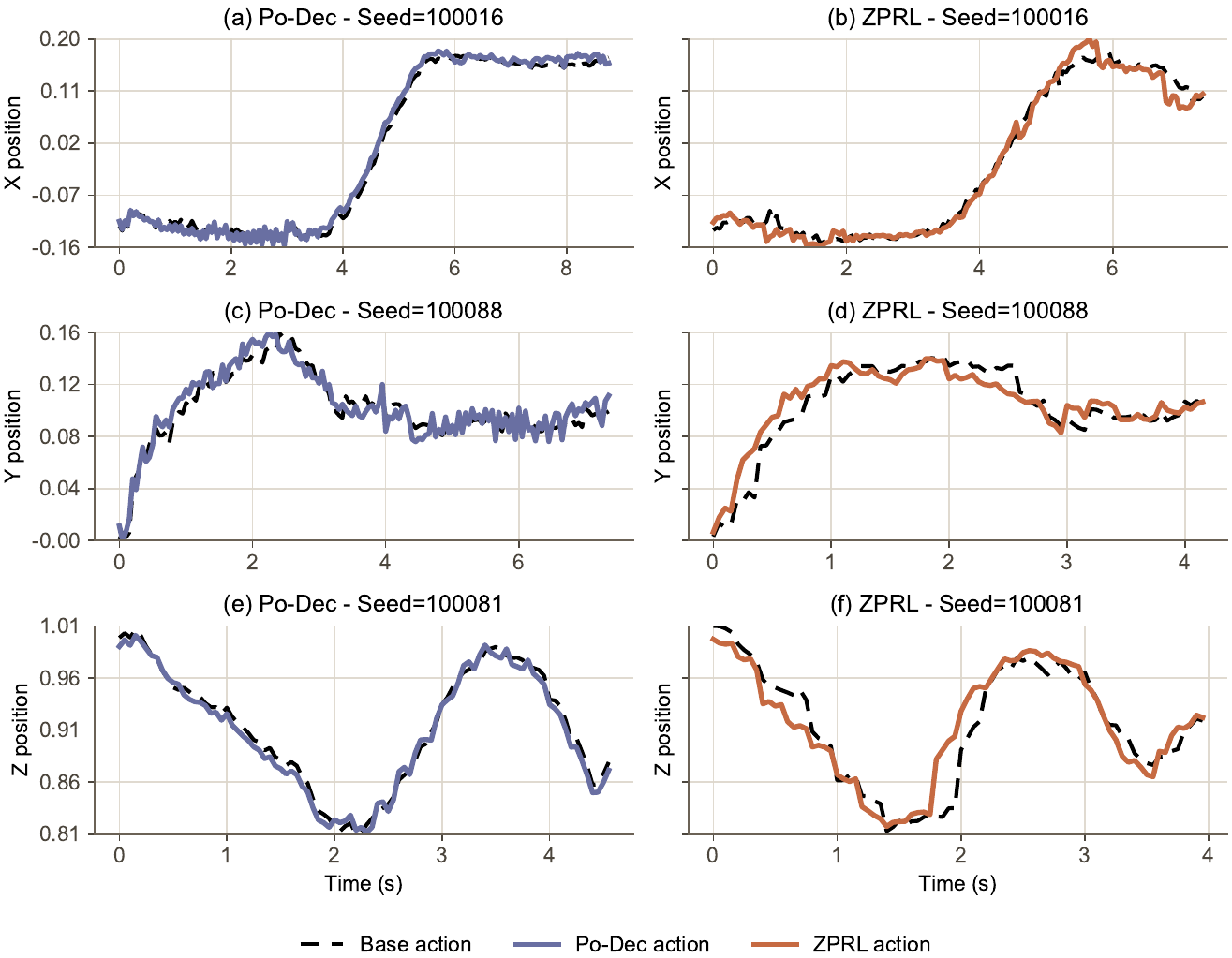}
    \par\vspace*{-3mm}
    \caption{\textbf{Qualitative comparison of action outputs on \texttt{Square}.}
    Three layouts are selected from the 150 evaluation layouts by the largest Po-Dec jerk variance per axis to characterize worst-cases. Both policies use 3.2M-step checkpoints. Rows show desired $x$-, $y$-, and $z$-position components, respectively, with Po-Dec on the left and ZPRL on the right. Dashed lines show the base-policy action. Each panel shows one rollout.}
    \vspace{-2mm}
    \label{fig:smooth_traj}
\end{figure}

Table~\ref{tab:smooth_metric} shows that ZPRL combines high success with smooth execution: it reaches 98.7\% SR with mean jerk of 19.7\,\si{m/s^3}, compared with 90.7\% and 32.3\,\si{m/s^3} for Po-Dec. EMA filtering reduces Po-Dec's jerk to 27.5\,\si{m/s^3}. Strong CAPS regularization~\cite{mysore2021caps} achieves slightly lower jerk than ZPRL, but reduces SR to 73.3\%. Thus, the benefit lies in the success-smoothness trade-off.
\Figref{fig:smooth_traj} illustrates the action outputs behind the observed-state smoothness comparison. In these cases, Po-Dec's desired-position commands show stronger oscillations, while ZPRL produces smoother corrections to the base action. Rollouts are shown separately because their unequal durations prevent direct temporal alignment.

These results show that bottleneck latent steering can combine high task success with smooth behavior without explicit temporal filtering or smoothness regularization.

\subsection{ZPRL in the Real World}
\label{sec:real}

\begin{figure*}[!t]
    \centering
    \includegraphics[width=0.85\textwidth]{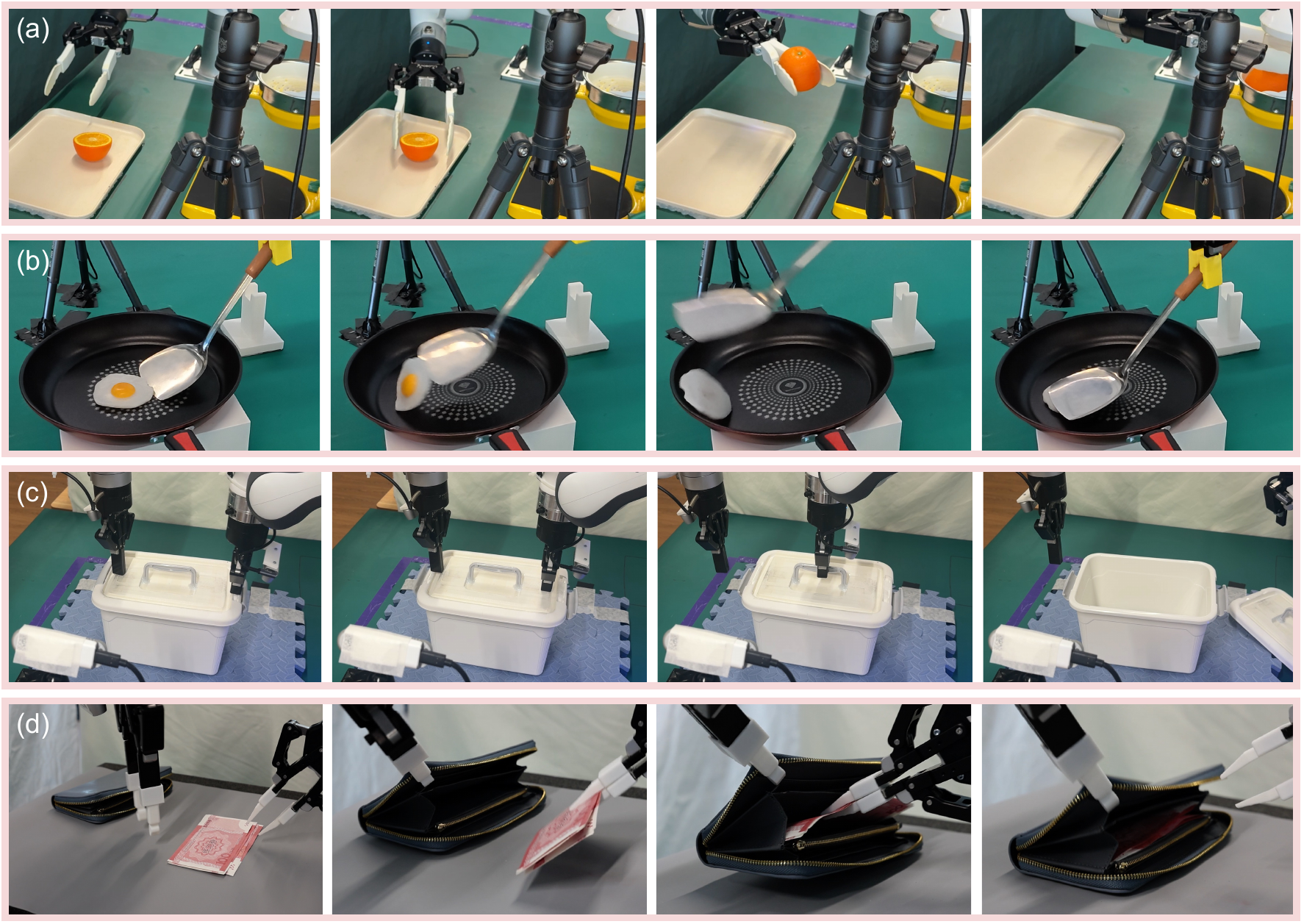}
    \vspace{-3mm}
    \caption{\textbf{Rollout trajectories for four real-world tasks.} Each row shows temporally ordered, subsampled snapshots from one rollout. From top to bottom: (a) \po, (b) \fe, (c) \ob, and (d) \ib.}
    \label{fig:real_trajs}
    \vspace{-3mm}
\end{figure*}

We further evaluate ZPRL on four challenging real-world manipulation tasks: \po, \fe, \ob, and \ib~(\Figref{fig:real_trajs}). These tasks cover a diverse range of physical skills, including single-arm pick-and-place, highly dynamic contact-rich manipulation, coordinated bimanual interaction, and deformable-object manipulation (DOM).
In \po, the robot must grasp one half of a real orange randomly placed on a tray (29\,\si{cm}\,$\times$\,20\,\si{cm}), rotate its wrist, and place the orange half onto the filter of a juicer. This task requires reliable visual localization, accurate grasping, and precise pose alignment at placement.
In \fe, the robot starts with a spatula grasped by a 3D-printed gripper and must insert it underneath a sunny-side-up egg model, flip the egg with a high end-effector (EE) acceleration of up to 6\,$\si{m/s^2}$, and drag it back to the center of the pan. This task is highly dynamic and sensitive to insertion depth and contact timing.
In \ob, two robot arms must coordinate to open the latches on both sides of a box, after which the right arm lifts and releases the lid. This task involves bimanual coordination, contact-rich interaction, and temporal synchronization between the two manipulators.
In \ib, two arms must cooperatively pick up four banknotes from the table, open a wallet, and insert the bills into the narrow inner pocket. This task requires bimanual coordination, DOM, and contact-rich insertion under partial occlusion.
Together, these tasks form a real-world benchmark that spans increasing levels of difficulty in dynamics, contact, and coordination, allowing us to assess whether the advantages of ZPRL remain beyond simulation.

\subsubsection{Training Base Policies}
For each task, we first collect human demonstrations with a Virtual Reality (VR) headset and use them to train the base flow policy. We adopt the same model architecture as in simulation, with slightly increased model capacity for real-world observations; detailed hyperparameters are provided in Appendix~\ref{app:arch}. The amount of offline data and collection time are summarized in Table~\ref{tab:real_data}.

\begin{table}[ht]
\caption{Amount of training data and collection time for each real-world task, reported as [offline $|$ online]. The number of online $(\vo,\va)$ pairs is not directly comparable to the offline count, because online action chunks are added to the replay buffer end-to-end, rather than as overlapping windows in the offline dataset.}
\label{tab:real_data}
\centering
\begin{tabular}{lccc}
\toprule
\textbf{Task} & \textbf{\# of trajs.} & \textbf{\# of $(\vo,\va)$} & \textbf{Time (h)} \\
\midrule
\po           & 100 $|$ ~370         & 29.1k $|$ ~6.6k      & 1.0 $|$ ~3.5       \\
\fe           & 150 $|$ 1350         & 20.2k $|$ 10.5k      & 2.0 $|$ ~8.5       \\
\ob           & ~82 $|$ ~330         & 23.5k $|$ ~5.5k      & 1.0 $|$ ~4.0       \\
\ib          & 100 $|$ ~949          & 76.8k $|$ 30.3k      & 2.5 $|$ 12.5       \\
\bottomrule
\end{tabular}
\par\vspace*{-2mm}
\end{table}

\begin{figure}[!t]
    \centering
    \includegraphics[width=0.80\columnwidth]{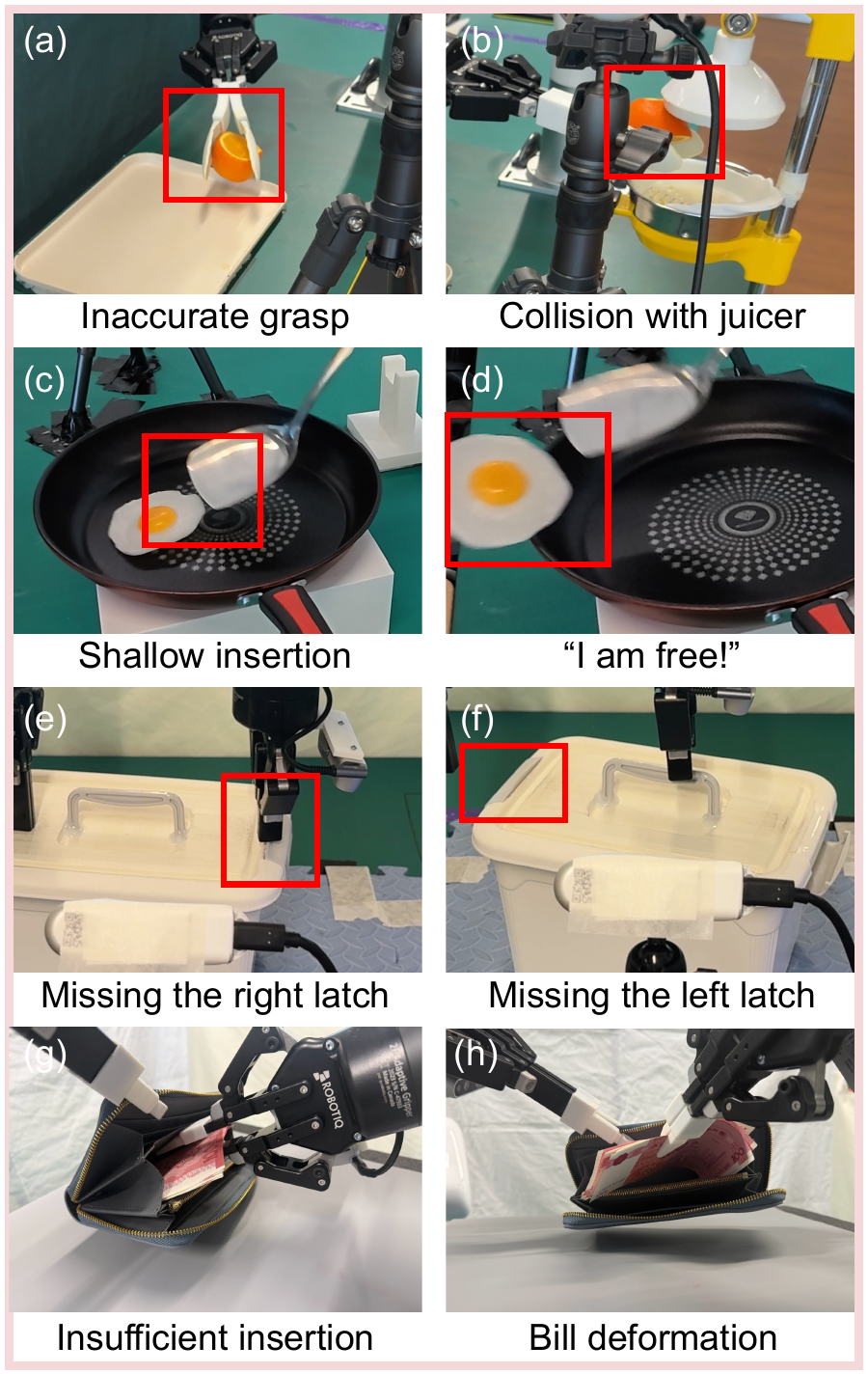}
    \vspace{-3mm}
    \caption{\textbf{Common failure modes in the real world.} In \po: (a) inaccurate grasping and (b) collision with the juicer. In \fe: (c) shallow insertion that fails to flip the egg and (d) overly high end-effector velocity causing the egg to fly out of the pan. In \ob: (e) missing the right or (f) the left latch. In \ib: (g) insertion stops halfway due to partial occlusion and (h) bills bend or collide with the wallet.}
    \vspace{-5mm}
    \label{fig:real_failure}
\end{figure}

The resulting base policies already achieve meaningful SRs, providing a reasonable starting point for online finetuning. However, they still exhibit several failure modes in real-world execution, as illustrated in \Figref{fig:real_failure}. In \po, failures come from inaccurate grasping and collisions with the juicer during placement. In \fe, shallow insertion may fail to lift the egg, while overly aggressive motion may throw the egg out of the pan. In \ob, the policy may miss either the left or the right latch, causing the opening to fail. In \ib, the bills may stop halfway during insertion due to partial occlusion, or bend and collide with the wallet edge. These remaining errors indicate that imitation alone does not fully resolve the fine-grained control and contact uncertainties in the real world, leaving clear room for improvement through online RL.

\subsubsection{Real-World Online RL}
For \po, \fe, and \ob, we finetune both Po-Dec and ZPRL from the same offline base checkpoint under the same interaction budget. For the more time-consuming \ib, we only run ZPRL, and use it as an additional real-world stress test for evaluating whether the proposed method remains deployable in a more complex bimanual DOM setting.
At the beginning of each episode, the task object is randomly initialized within a predefined region of the workspace: the tray in \po; half of the pan in \fe; a 35\,cm\,$\times$\,20\,cm rectangular area with an additional rotation range within $\pm 10^\circ$ in \ob. For \ib, the banknotes are placed below the right arm with loose alignment and a lateral displacement within $\pm 3$\,cm, while the wallet is placed in the upper-left region of the table with its horizontal position randomized within $\pm 5$\,cm and its orientation randomized within $\pm 10^\circ$.
The policy takes image observations together with robot joint angles as input, and outputs a chunk of desired EE poses for the next few timesteps to low-level controllers.
A human supervisor monitors the rollout, resets the workspace between episodes, and provides the final sparse reward. Specifically, we use $r_t=1$ if and only if the task is successfully completed, and $r_t=0$ otherwise. Each episode terminates upon one of three conditions: (i) task success; (ii) an irrecoverable failure; or (iii) reaching a predefined maximum horizon.
We use a synchronous interaction-and-update loop: model updates occur only between episodes, so controller execution is not interrupted. We use $\mathrm{UTD}=5$ for \ib~due to its complexity and $\mathrm{UTD}=2$ for other tasks. Inference latency and update throughput can be found in Appendix~\ref{app:runtime}.

\begin{figure}[!t]
    \centering
    \includegraphics[width=0.475\columnwidth]{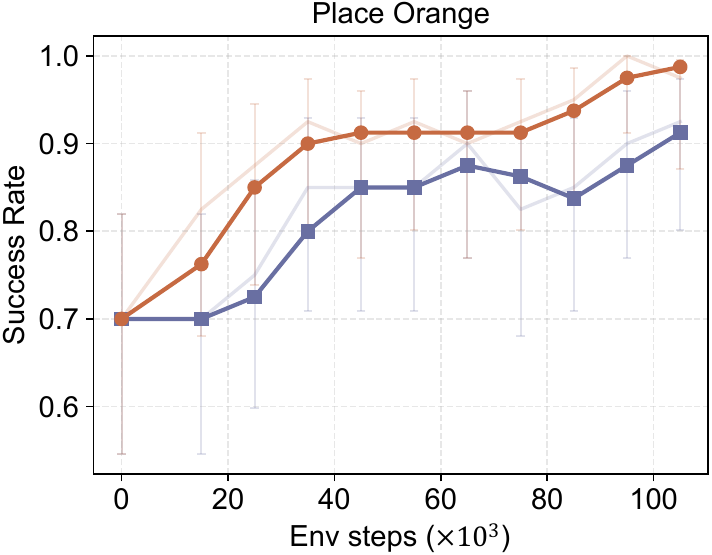}\hfill
    \includegraphics[width=0.475\columnwidth]{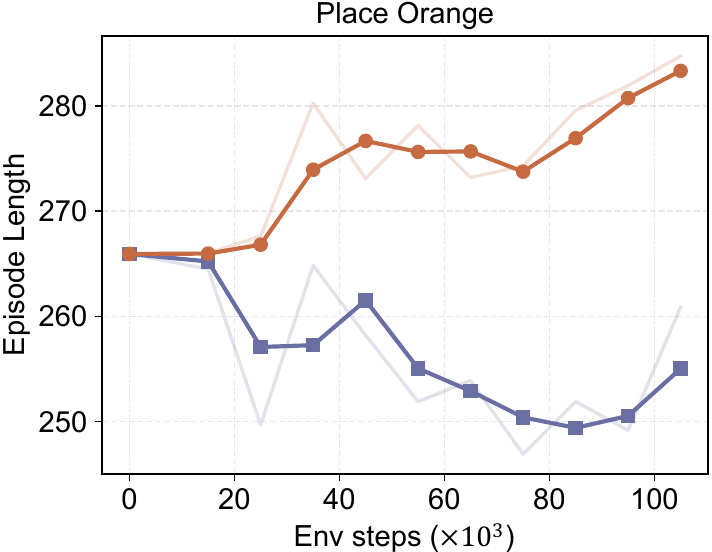}\\[2mm]
    \includegraphics[width=0.475\columnwidth]{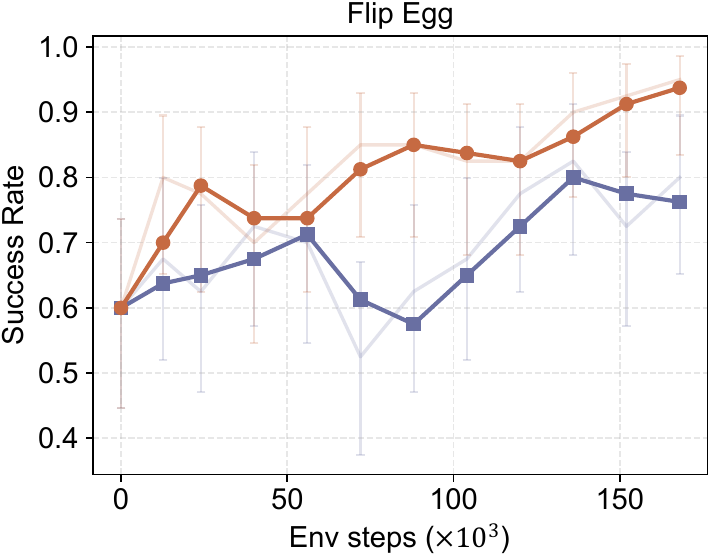}\hfill
    \includegraphics[width=0.475\columnwidth]{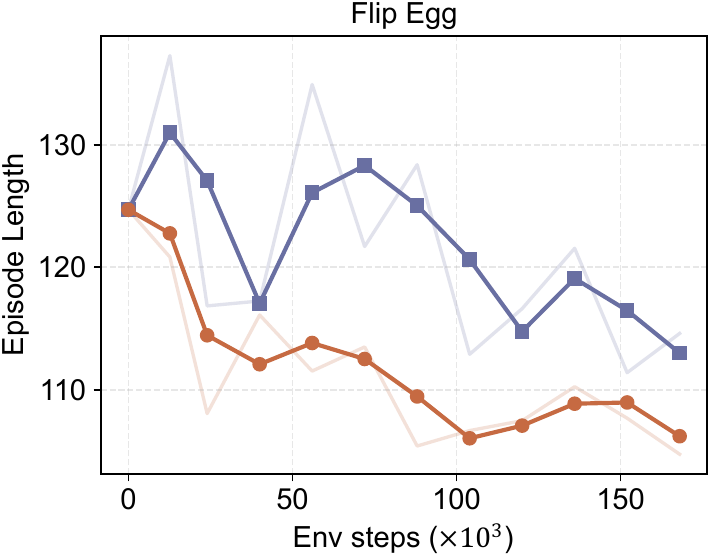}\\[2mm]
    \includegraphics[width=0.475\columnwidth]{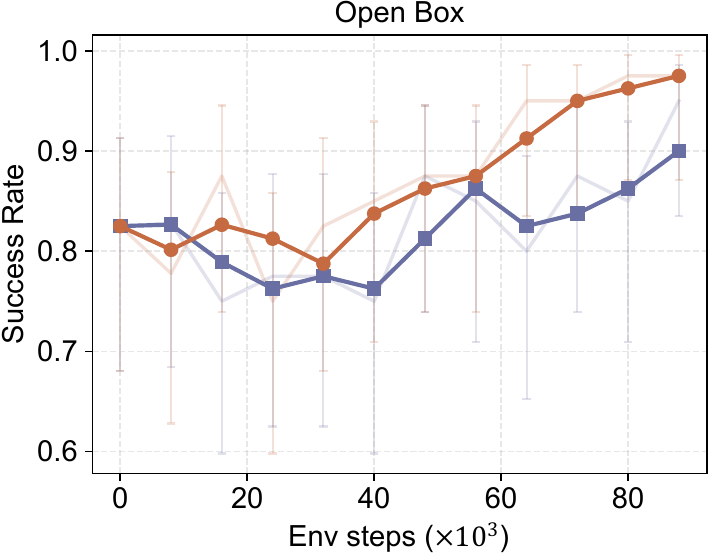}\hfill
    \includegraphics[width=0.475\columnwidth]{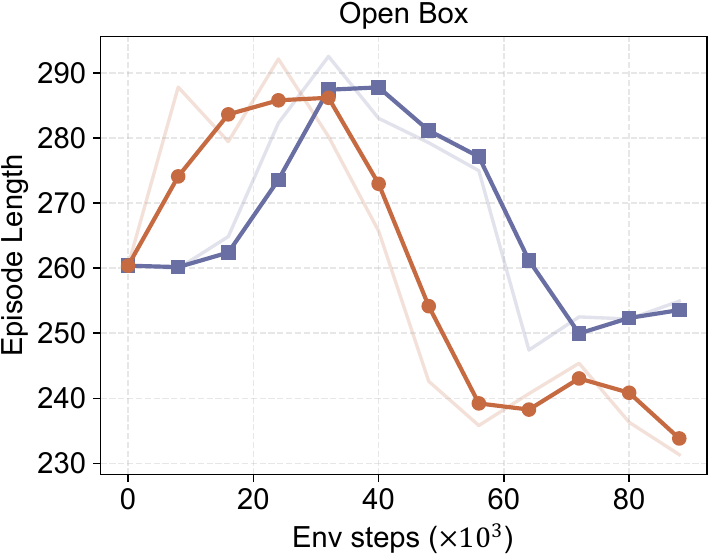}\\[2mm]
    \includegraphics[width=0.475\columnwidth]{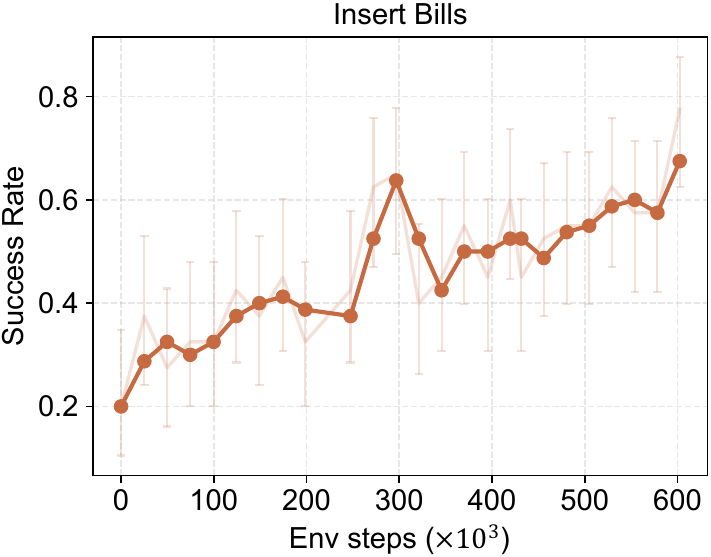}\hfill
    \includegraphics[width=0.475\columnwidth]{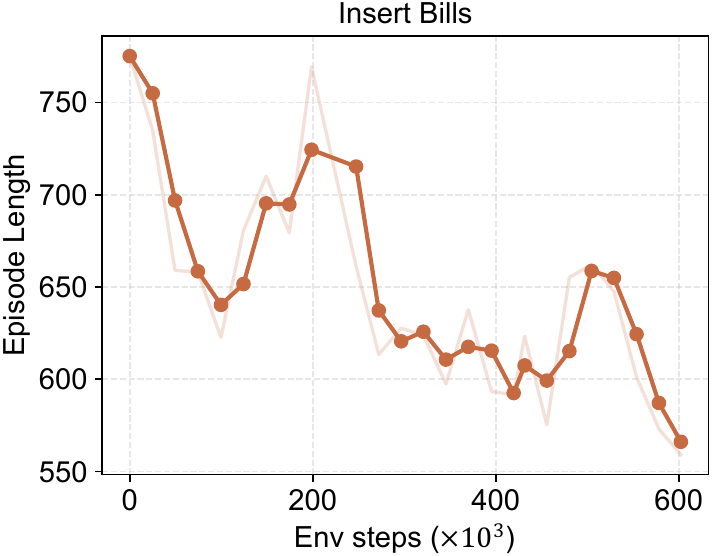}
    \vspace{2pt}
    \includegraphics[width=0.4\columnwidth]{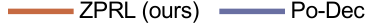}
    \vspace{-3mm}
    \caption{\textbf{Real-world RL curves on four tasks.} SR (left) and average episode length (right) are plotted against environment steps. Each point comes from a window of 40 training episodes. Light curves show these values and dark curves show two-point moving averages. Bars are pointwise 95\% Wilson intervals for the 40 episodes in the window, which may overlap between adjacent points. For \ib, only ZPRL is trained due to the substantially higher hardware and time cost.}
    \vspace{-5mm}
    \label{fig:real_main}
\end{figure}

\subsubsection{Main Results}
We conduct one online training run per task and method. To avoid pausing training for separate evaluation frequently, we report the SR over a rolling window of the most recent 40 episodes after updating starts, so each point reflects performance during learning and carries exploration noise except the starting point from the base. \Figref{fig:real_main} plots SR and episode length against environment steps, both computed over the same window and including failures and timeouts.

On the three tasks where both methods are evaluated (\po, \fe, and \ob), ZPRL reaches high SR faster than Po-Dec and achieves higher final SR, with larger gains on the more challenging \fe~(by roughly 12.5 pp) and \ob~(by about 7.5 pp).
Reaching high SR with fewer environment steps supports the interaction-efficiency benefit of bottleneck latent steering. Episode lengths vary across tasks and methods, with longer ZPRL episodes on \po. Because these averages mix successful and failed trials, they do not directly measure execution efficiency.

On the additional \ib, where only ZPRL is trained online due to the high hardware cost, ZPRL substantially improves the base policy from a low SR (20\%) to a final SR of 77.5\%. This result suggests that the proposed method can also be deployed in complex tasks involving fine-grained manipulation, deformable objects, and bimanual coordination, and can improve itself through interaction.

\subsubsection{Robustness of ZPRL Policies}
\begin{figure}[!t]
    \centering
    \includegraphics[width=0.9\columnwidth]{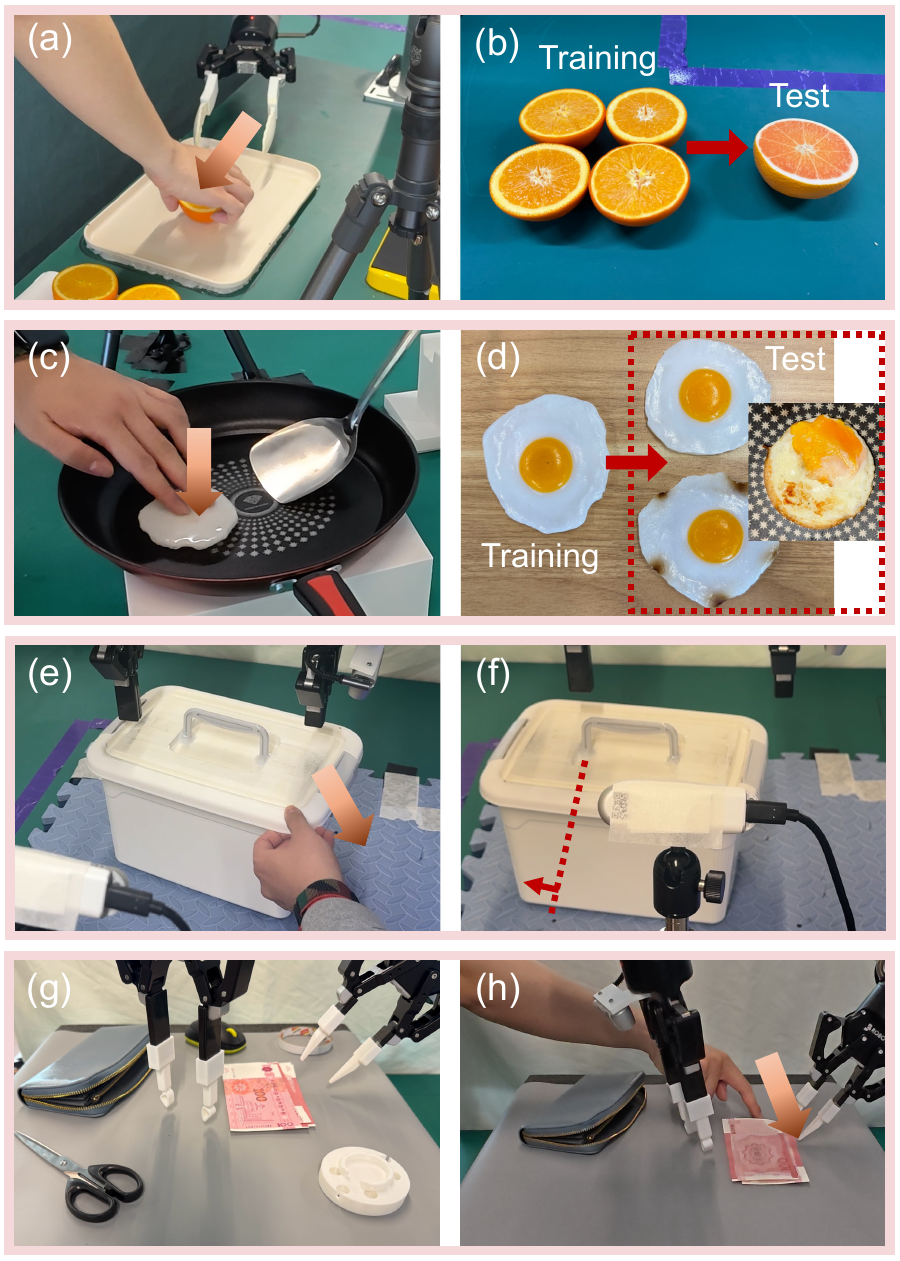}
    \par\vspace*{-3mm}
    \caption{\textbf{Robustness test cases for evaluating ZPRL under different disturbances.} In (a), (c), (e), and (g), a human perturbs the object after the episode begins. In (b) and (d), the training object is replaced with a novel one. In (f), the initial object pose is perturbed with additional positional or rotational offsets. In (h), several distractors are placed on the workspace.}
    \vspace{-3mm}
    \label{fig:real_robust_vis}
\end{figure}

We further evaluate the final ZPRL policies under unseen deployment disturbances, introduced only at evaluation time with no policy updates. We design task-specific perturbations, including human intervention after the episode starts, object replacement, and out-of-distribution (OOD) initial layouts. These test cases are illustrated in \Figref{fig:real_robust_vis}. We roll out the policy 10 times for each case, and summarize the zero-shot SRs in Table~\ref{tab:real_robust_res}.

\begin{table}[!ht]
\caption{Success rate (SR) under human disturbances, novel objects, and OOD initial layouts. All policies are evaluated zero-shot after online finetuning.}
\label{tab:real_robust_res}
\centering
\begin{tabular}{clc}
\toprule
\textbf{Task}        & \multicolumn{1}{c}{\textbf{Test Case}} & \textbf{SR} \\ \midrule
\multirow{2}{*}{\po} & Human disturbance                      & 0.7         \\
                     & Model orange                           & 1.0         \\ \midrule
\multirow{4}{*}{\fe} & Human disturbance                      & 0.8         \\
                     & Different shape                        & 0.8         \\
                     & Different color                        & 0.9         \\
                     & Real egg                               & 0.5         \\ \midrule
\multirow{2}{*}{\ob} & Human disturbance                      & 0.6         \\
                     & OOD position                           & 0.7         \\ \midrule
\multirow{2}{*}{\ib} & Visual distractors                     & 0.4         \\
                     & Human disturbance                      & 0.5         \\ \midrule
Average              &                                        & 0.69        \\ \bottomrule
\end{tabular}
\vspace{-1mm}
\end{table}

Overall, ZPRL achieves an average SR of 69\% across the tested disturbances, supporting moderate robustness of the base-plus-residual policy under these variations. In particular, the policies remain reasonably robust to direct human disturbances, with SRs of 0.7 on \po, 0.8 on \fe, and 0.6 on \ob. They also generalize well to moderate appearance changes: the \po\ policy achieves 1.0 SR on the visually different plastic orange model, while the \fe\ policy achieves 0.8 and 0.9 SR on eggs with different shape and color, respectively. For \ob, the policy still achieves 0.7 SR when the box is initialized with an additional positional (2\,cm) or rotational (10$^\circ$) offset, suggesting tolerance to mild shifts.

A challenging case is the \textit{Real egg} setting in \fe, where SR drops to 0.5. Besides color or shape changes, replacing the rigid model with a much softer real egg changes the contact mechanics of insertion and flipping. In this case, a strategy that only inserts the spatula by a few centimeters, which is sufficient for the model egg, is no longer reliable.

The policy shows lower robustness on disturbed \ib~because this task is highly sensitive to both visual and physical perturbations. The thin and deformable paper bills can easily bend or become misaligned after disturbance, which makes fine-grained insertion and recovery difficult. Since such recovery behaviors are not well covered in the offline dataset, SRs drop substantially under these conditions.
These results suggest that VIB latents can suppress task-irrelevant information to some degree but do not guarantee invariance to such variations.

\subsubsection{Smoothness of ZPRL Actions}
We evaluate the final Po-Dec and ZPRL policies on \po with 20 deterministic, unfiltered rollouts each. At similar mean velocity (0.09\,\si{m/s}), ZPRL reduces translational acceleration from 1.8 to 0.9\,\si{m/s^2} and jerk from 88.0 to 46.3\,\si{m/s^3}; the full statistics and protocol are given in Appendix~\ref{app:real_smooth}.
\Figref{fig:real_smooth} provides a separate qualitative view of exploration at middle-stage checkpoints. Both methods start from the same initial state near the edge of the plate, with exploration enabled. ZPRL follows a more coherent path, while Po-Dec exhibits more frequent oscillations.
On hardware, only Po-Dec on \fe uses a three-step pose EMA filter ($\omega=0.7$) to mitigate frequent failure caused by oscillated motions. All other cases are unfiltered.

\section{Limitations and Future Directions}
\label{sec:limitation}

Despite the asymptotic SR and learning efficiency of ZPRL, the current framework still has several limitations that lead us to promising directions for future work.

\noindent\textbf{Dependence on the Base Policy.}
ZPRL is not designed to synthesize absent behaviors while the base policy remains frozen, or to handle deployment far outside the offline data support. If a required primitive is missing, such as flattening bent bills in \ib, targeted demonstrations and an update to the base are required first. We recommend constructing and validating the behavioral coverage sufficiently, then freezing the base for a single ZPRL phase. Offline imitation establishes behavioral coverage while online ZPRL maximizes deployment return by improving how existing behaviors are used. These are complementary stages.

\begin{figure}[!t]
    \centering
    \includegraphics[width=0.49\columnwidth]{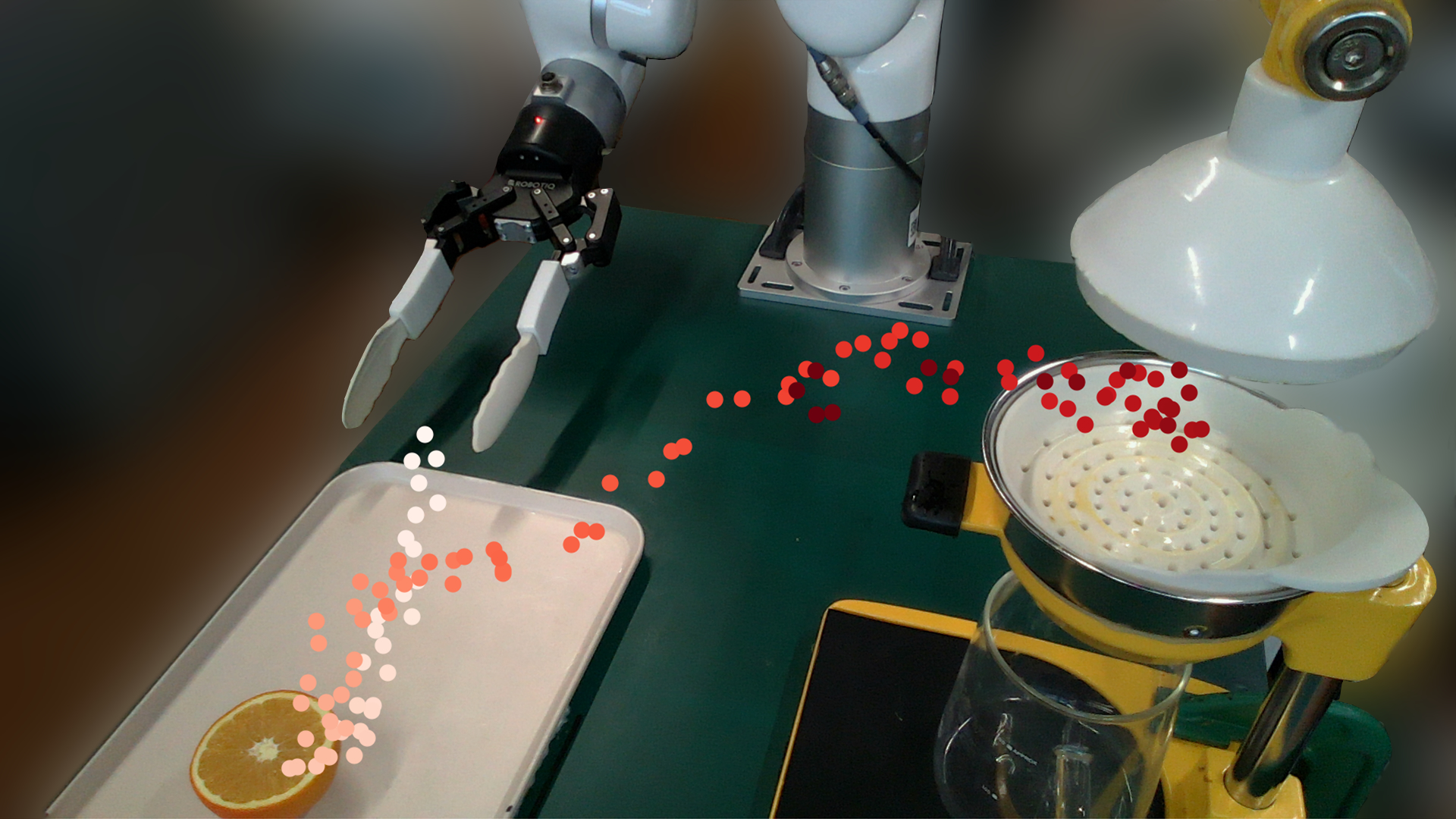}\hfill
    \includegraphics[width=0.49\columnwidth]{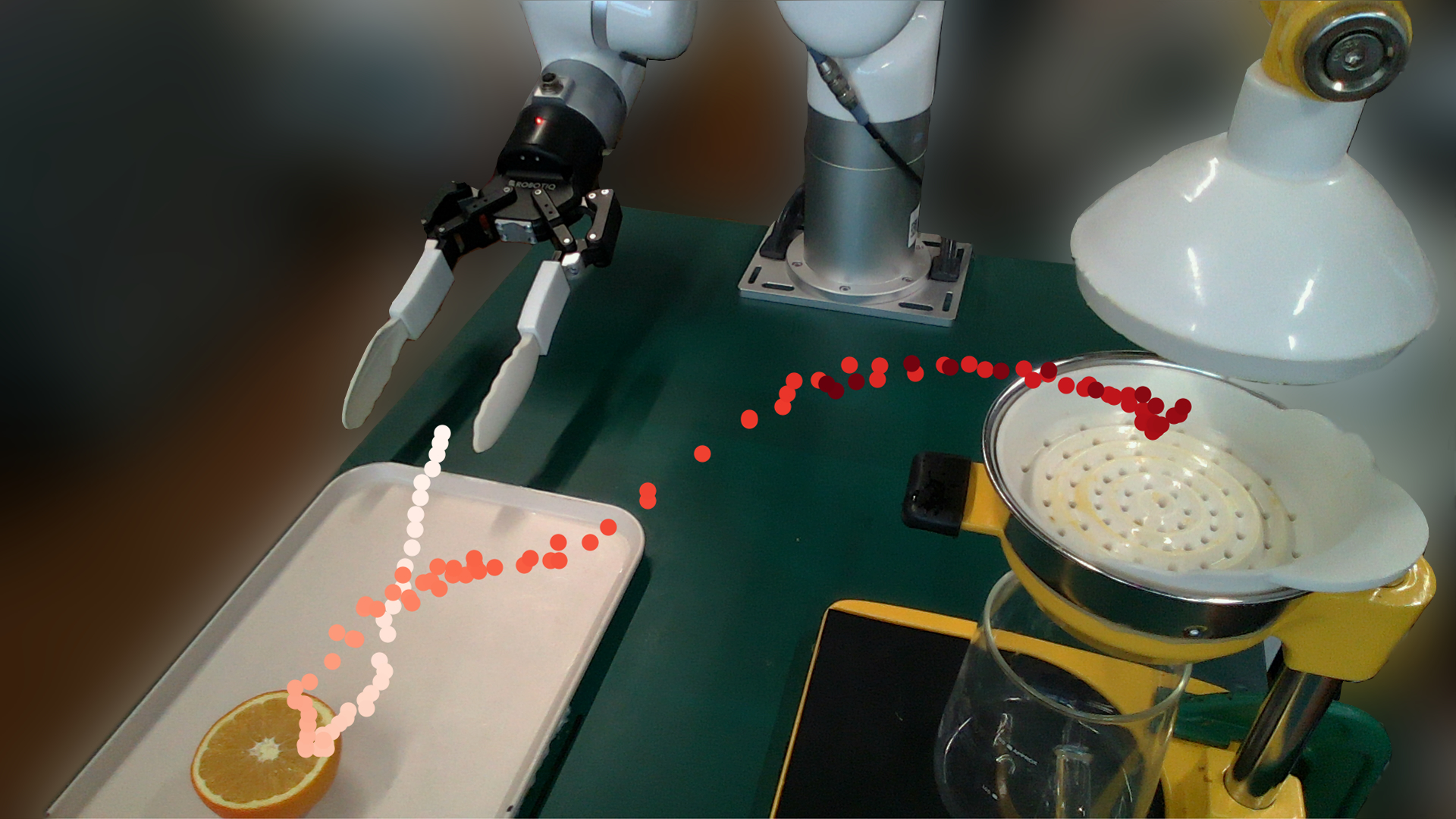}
    \vspace{-2mm}
    \caption{\textbf{Qualitative exploratory trajectories on \po.} Po-Dec (left) and ZPRL (right) use middle-stage checkpoints with exploration enabled, starting near the edge of the plate. Dots show recorded EE positions projected onto the image; darker dots indicate later timesteps.}
    \vspace{-3mm}
    \label{fig:real_smooth}
\end{figure}

\noindent\textbf{Post-hoc VIB and Data Requirements.}
Post-hoc and jointly trained VIB interfaces yield closely matched online performance on \texttt{Transport} and \texttt{Box-close} (Appendix~\ref{app:posthoc}). On \texttt{Transport}, fitting the VIB on proficient-human (PH) data after training the base on MH data also supports effective online adaptation in spite of degraded base performance, showing that access to the original training dataset is unnecessary. However, attachment still requires accessible intermediate features and ``task-relevant'' offline data. This same-task data shift does not guarantee arbitrary cross-task transfer.

\noindent\textbf{Extension to Broader Policy Architectures.}
Our insertion point is the observation-conditioning boundary, not an arbitrary hidden layer. This assumes an accessible interface, as in encoder-action-expert policies and relatively decoupled VLAs~\cite{starvla}, but does not directly cover conditioning distributed across KV caches in multiple layers. For those architectures, two possible approaches are: (a) compressing vision-language tokens into a compact state, as in RL Token~\cite{xu2026rl}, and applying a second compression to obtain the bottleneck; (b) when dense visual features must be maintained, learning a steering token and adding it to those features. Both are design proposals that require scale-up experiments in future work.

\noindent\textbf{Beyond Reinforcement Learning.}
Finally, although this work focuses on online RL, the idea of policy steering through a bottleneck latent may extend beyond RL. Prior work on iterative IL with VIB has shown that collecting additional data over multiple rounds can improve policy quality~\cite{jin2025soe}, but such methods typically retrain or finetune the entire policy after each data collection round. It is worth studying whether latent-space steering can serve as a lighter-weight alternative when the offline dataset is already large and each new round contributes only a relatively small amount of additional data. In that regime, steering may offer a more targeted way to adapt to long-tail or failure-prone cases, rather than repeatedly mixing all collected samples into a single growing dataset.

\section{Conclusion}

We presented Z-Perturbation Reinforcement Learning (ZPRL), a lightweight post-training framework that improves pretrained robot manipulation policies within their behavioral support through online RL.
Building on SOE's VIB-based policy interface, ZPRL demonstrates that reward-optimized residual steering of this latent provides an effective alternative to weight-, action-, and diffusion-noise-space adaptation.
Across simulation and real-world tasks, ZPRL improves sample efficiency with competitive final performance against strong post-training baselines. It also achieves lower acceleration and jerk than the baselines while maintaining task success, without explicit temporal filtering or smoothness regularization. These results suggest that the choice of steering interface is a key factor in RL post-training, and that a compact, semantically organized latent offers a practical middle ground between full finetuning and direct action-space correction.



\section*{Acknowledgment}
The authors used Gemini CLI and OpenAI models, including GPT-5.6 and GPT-6, to assist with writing and debugging code for the experiments in Section~V and with improving the clarity of the manuscript. Claude Code was used for proofreading. The authors reviewed all AI-generated content and take full responsibility for the final manuscript.

\bibliographystyle{IEEEtran}
\bibliography{IEEEabrv, ref_abrv}

\input{appendix}

\end{document}

%% file: math_symbols.tex
\usepackage{amsmath,amsfonts,bm}

\def\Figref#1{Fig.~\ref{#1}}

\def\Secref#1{Section~\ref{#1}}

\def\eqref#1{equation~(\ref{#1})}
\def\Eqref#1{Equation~(\ref{#1})}

\def\1{\bm{1}}

\def\rva{{\mathbf{a}}}

\def\rvc{{\mathbf{c}}}

\def\rvz{{\mathbf{z}}}

\def\va{{\bm{a}}}

\def\vc{{\bm{c}}}

\def\vo{{\bm{o}}}

\def\vq{{\bm{q}}}

\def\vs{{\bm{s}}}

\def\vw{{\bm{w}}}

\def\vz{{\bm{z}}}

\def\mI{{\bm{I}}}

\DeclareMathAlphabet{\mathsfit}{\encodingdefault}{\sfdefault}{m}{sl}
\SetMathAlphabet{\mathsfit}{bold}{\encodingdefault}{\sfdefault}{bx}{n}

\def\gA{{\mathcal{A}}}

\def\gD{{\mathcal{D}}}
\def\gE{{\mathcal{E}}}

\def\gL{{\mathcal{L}}}
\def\gM{{\mathcal{M}}}
\def\gN{{\mathcal{N}}}
\def\gO{{\mathcal{O}}}

\def\gS{{\mathcal{S}}}

\newcommand{\E}{\mathbb{E}}



%% file: appendix.tex
\appendix
\setcounter{table}{0}
\setcounter{figure}{0}
\renewcommand{\thetable}{A.\Roman{table}}
\renewcommand{\thefigure}{A.\arabic{figure}}
\renewcommand{\theHtable}{appendix.\arabic{table}}
\renewcommand{\theHfigure}{appendix.\arabic{figure}}

\subsection{Additional Experimental Details}\label{app:details}

\subsubsection{Task Selection}\label{app:selection}
Our task selection balances sufficient room for online improvement, skill-type coverage, and the availability of offline datasets and baseline results. Among omitted tasks, \texttt{Robomimic\_lift} overlaps with the pick-and-place behavior in \texttt{Can}, while \texttt{Adroit\_relocate} shares grasp-and-move skills with \texttt{Hammer}. Our available \texttt{Metaworld} dataset contains four tasks, but the base policies already achieve 96\% success rate (SR) on \texttt{Peg-Insert-Side} and 82\% on \texttt{Disassemble}, leaving limited margin for distinguishing online methods. We use \texttt{Square} for controlled ablations because its base SR is about 40\%, leaving room for online improvement. \texttt{Robomimic\_tool\_hang} is included in supplementary analyses.

\subsubsection{Model Architectures and Training}\label{app:arch}
Our implementation is built on top of Diffusion Policy~\cite{chi2023diffusion}, but replaces the diffusion objective with the flow-matching objective and adds the online RL pipeline. The policy architectures in simulation and in the real world share the same overall design. The observation representation consists of two parts: (1) an image embedding from each camera at each observation step, with embedding dimension 256, produced by a ResNet-18 encoder trained from scratch followed by a dimension-reduction projection; and (2) a raw proprioceptive vector, namely end-effector poses in simulation and joint angles in the real world.

The flow policy backbone is a 1D U-Net with convolutions applied along the action-chunk dimension. The only architecture difference between simulation and real-world experiments is the channel width of the 1D U-Net, which is 128-256-512 in simulation and 256-384-512 in the real world. The added VIB module consists of two multi-layer perceptrons (MLPs), namely the encoder and decoder, each with four layers of width 256 and GELU activations. During offline training, we discretize the flow schedule into 100 steps, \ie $1, 0.99, 0.98, \dots, 0.01$. The flow denoising steps during inference are included in Table.~\ref{tab:env_sim} and~\ref{tab:env_real}. Note that we use 20 for \texttt{Robomimic\_tool\_hang} due to its task complexity. Offline pretraining lasts for 1,000 epochs with batch size 128 or 256 depending on the dataset size.

\begin{table}[!b]
\setlength{\tabcolsep}{1.9pt}
\caption{\textbf{Hyperparameters of the online RL policy.} Values are reported as [simulation $|$ real world] when they differ. Task-specific values are listed in parentheses. All values are shared between ZPRL and Po-Dec.}
\label{tab:param_online}
\centering
\begin{tabular}{lc}
\toprule
\textbf{Hyperparameters}         & \textbf{Value}             \\ \midrule
Batch size                       & 256                        \\
Actor learning rate              & 1e-4                       \\
Critic learning rate             & 3e-4                       \\
Discount factor $\gamma$         & 0.99 (\texttt{Transport} 0.997, \ib~0.998) \\
Optimizer                        & Adam                       \\
UTD                              & 1 $|$ 2 (\ib 5)            \\
Target entropy                   & $-\dim(\va)$/2             \\
Target update rate ($\tau$)      & 0.005                      \\
Initial temperature              & 0.01                       \\
Hidden size                      & 256                        \\
\# of layers                     & 4                          \\
Activation                  & GELU (\texttt{MW, Adroit} add layer norm)\\
\# of Q functions                & 2 (\texttt{door} 5) $|$ 5 (\ib 10) \\
Actor parameterization           & $\tanh(\mathrm{Gaussian})$ \\
Actor logstd max                 & 2                          \\
Actor logstd min                 & -10                        \\ \bottomrule
\end{tabular}
\vspace{-3mm}
\end{table}

\begin{table}[!t]
\caption{\textbf{Task-specific hyperparameters for ZPRL in simulation.} Here $\dim(\vz)$ is the bottleneck latent dimension, $\lambda$ is the perturbation scale, $\dim(\vq)$ is the proprioceptive dimension, and $\dim(\va)$ is the action dimension per step. Environment steps are the number of predicted chunks multiplied by the action chunk size (further multiplied by 2 for action repeat in \texttt{Adroit}). During warm-up, a randomly initialized policy fills the replay buffer without performing any gradient update.}
\label{tab:env_sim}
\centering
\setlength{\tabcolsep}{5.5pt}
\begin{tabular}{lcccc}
\toprule
\textbf{Hyperparameters} & \textbf{Can}  & \textbf{Square} & \textbf{Transport} & \textbf{Box-close} \\ \midrule
$\dim(\vz)$          & 16           & 16           & 32           & 8     \\
$\lambda$            & 0.25         & 0.2          & 0.1          & 1.25  \\
Image size           & 84$\times$84 & 84$\times$84 & 84$\times$84 & 84$\times$84 \\
Crop size            & 76$\times$76 & 76$\times$76 & 76$\times$76 & 80$\times$80 \\
\# of cameras        & 2            & 2            & 4            & 1            \\
$\dim(\vq)$          & 9            & 9            & 18           & 9            \\
Obs chunk size       & 1            & 1            & 1            & 1            \\
$\dim(\va)$          & 10           & 10           & 20           & 4            \\
Action chunk size    & 4            & 4            & 5            & 2            \\
Action repeat        & 1            & 1            & 1            & 1            \\
\# of warm-up chunks & 1000         & 2000         & 4000         & 10000 \\
Max env steps per ep. & 400 & 400 & 700 & 200 \\
Flow steps    & 2   & 2 & 2 & 5 \\
\midrule
\textbf{Hyperparameters} & \textbf{Door} & \textbf{Hammer} & \textbf{Pen}       & \textbf{Push-wall} \\
\midrule
$\dim(\vz)$          & 16           & 16           & 16           & 8     \\
$\lambda$            & 0.75         & 0.5          & 0.4          & 0.5   \\
Image size           & 84$\times$84 & 84$\times$84 & 84$\times$84 & 84$\times$84 \\
Crop size            & 80$\times$80 & 80$\times$80 & 80$\times$80 & 80$\times$80 \\
\# of cameras        & 1            & 1            & 1            & 1            \\
$\dim(\vq)$          & 24           & 24           & 24           & 9            \\
Obs chunk size       & 1            & 1            & 1            & 1            \\
$\dim(\va)$          & 28           & 26           & 24           & 4            \\
Action chunk size    & 2            & 2            & 2            & 2            \\
Action repeat        & 2            & 2            & 2            & 1            \\
\# of warm-up chunks & 4000 & 4000 & 2000 & 10000 \\
Max env steps per ep. & 200 & 200 & 150 & 200 \\
Flow steps    & 2   & 2 & 2 & 5 \\
\bottomrule
\end{tabular}
\vspace{-1mm}
\end{table}

\begin{table}[!t]
\caption{\textbf{Task-specific hyperparameters for ZPRL in the real world.}}
\label{tab:env_real}
\centering
\begin{tabular}{lcc}
\toprule
\textbf{Hyperparameters} & \textbf{\po} & \textbf{\fe} \\
\midrule
$\dim(\vz)$              & 32             & 32             \\
$\lambda$                & 0.2            & 0.2            \\
Image size               & 110$\times$280 & 128$\times$128 \\
Crop size                & 106$\times$276 & 124$\times$124 \\
\# of cameras            & 1              & 1              \\
$\dim(\vq)$              & 7              & 7              \\
Obs chunk size           & 1              & 3              \\
$\dim(\va)$              & 7              & 6              \\
Action chunk size        & 16             & 16             \\
Env frequency (Hz)       & 30             & 10             \\
\# of warm-up chunks     & 300            & 300            \\
Max env steps per ep.     & 500     & 250     \\
Flow steps        & 2       & 2       \\
\midrule
\textbf{Hyperparameters} & \textbf{\ob} & \textbf{\ib} \\
\midrule
$\dim(\vz)$              & 32             & 32             \\
$\lambda$                & 0.175          & 0.15           \\
Image size               & 128$\times$128 & 120$\times$160 \\
Crop size                & 124$\times$124 & 116$\times$156 \\
\# of cameras            & 3              & 3              \\
$\dim(\vq)$              & 16             & 16             \\
Obs chunk size           & 1              & 2              \\
$\dim(\va)$              & 16             & 16             \\
Action chunk size        & 16             & 24             \\
Env frequency (Hz)       & 20             & 20             \\
\# of warm-up chunks     & 500            & 500            \\
Max env steps per ep.     & 400     & 1000    \\
Flow steps        & 2       & 10       \\
\bottomrule
\end{tabular}
\vspace{-3mm}
\end{table}

The online policy $\pi_\mathrm{on}$ uses the same architecture in both simulation and real-world experiments. Its hyperparameters are summarized in Table~\ref{tab:param_online}. \texttt{Adroit} tasks are one exception, where we add layer normalization to the critic networks to stabilize training under dense rewards.
Tables~\ref{tab:env_sim} and~\ref{tab:env_real} summarize the task-specific settings for the simulation and real-world experiments, respectively.

\textbf{Evaluation protocol.} On \texttt{Can}, \texttt{Door}, \texttt{Hammer}, \texttt{Pen}, \texttt{Box-close}, and \texttt{Push-wall}, we periodically evaluate each checkpoint in separate parallel environments with stochasticity disabled, using 50 episodes on \texttt{Robomimic} and 100 on \texttt{Adroit} and \texttt{Metaworld}. \texttt{Square}, \texttt{Transport}, and \texttt{Tool\_hang} are GPU memory-intensive, so parallel evaluation environments would slow training. For these tasks we instead report the training-time SR over the most recent 100 completed episodes. This metric includes exploration noise and lags behind the current policy, so it may underestimate performance. All methods on the same task use the identical protocol, and Table~\ref{tab:train_stats} reports training-time metrics. In the real world, the learning curves similarly report SR over a rolling window of the most recent 40 episodes rather than separate evaluation runs. Base policies, robustness tests, and the smoothness evaluation use separate deterministic rollouts. Both methods on a task use the same window size.

\subsubsection{Practical Guideline for Choosing \texorpdfstring{$\lambda$}{lambda}}\label{app:lambda}
A practical question is how large the latent perturbation should be. Intuitively, $\lambda$ should be chosen to ensure that the online residual has substantial influence on the bottleneck to steer the base policy. If $\lambda$ is too large, however, the perturbed latent may move outside the local support of the pretrained latent distribution, making the decoded feature less reliable and degrading the frozen action prior, as shown in \Secref{app:whatzprl}.
In practice, we find it useful to choose $\lambda$ according to the ratio between the magnitude of perturbation $\Delta\vz$ and the original $\vz$. A convenient starting point is to set $\lambda$ such that $\mathrm{RMS}(\lambda \Delta \vz)$ is roughly 10\%--20\% of $\mathrm{RMS}(\vz)$ (RMS, root-mean-square), and then adjust it according to early online learning behavior: increasing $\lambda$ when learning is stable but slow, and decreasing it when performance degrades quickly. Final task-specific values are listed in Tables~\ref{tab:env_sim} and~\ref{tab:env_real}.

\subsubsection{Baseline Implementations}\label{app:baselines}

\textbf{DSRL.} We reproduce the DSRL-SAC version in the original paper. It steers the frozen base policy by learning input noise. Following the official implementation, we adopt noise repeating: we predict noise for one action step on \texttt{Transport}, \texttt{Push-wall}, and all \texttt{Adroit} tasks, and two steps elsewhere, then repeat it across the remaining action chunk. We use an ensemble of 10 Q functions on \texttt{Push-wall} and 5 elsewhere. Training bootstraps at timeouts, includes entropy in the Q targets, and applies layer normalization to both actor and critic networks.

\textbf{Policy Decorator (Po-Dec).} Po-Dec uses bounded residuals with per-task scales and progressive exploration durations listed in Table~\ref{tab:podec_schedule_scale}. All other SAC hyperparameters follow ZPRL's settings in Table~\ref{tab:param_online}. The residual scale multiplies the $\tanh$-bounded actor output. Main simulation evaluations are deterministic. On hardware, a three-step desired-pose exponentially weighted moving average (EMA) filter is applied only to Po-Dec on \fe because its unfiltered actions are too jittery to complete this precision-critical task:
\[
\hat \va_t=\frac{\sum_{i=0}^{2}\omega^i \va_{t-i}}{\sum_{i=0}^{2}\omega^i},\qquad \omega=0.7.
\]
Quaternion components are normalized after averaging. No other hardware case uses a filter.

\textbf{DPPO and ReinFlow.} Results are extracted from the raw training logs released with ReinFlow~\cite{zhang2025reinflow} for \texttt{Robomimic} and RL-100~\cite{lei2025rl100} for \texttt{Adroit} and \texttt{Metaworld} because offline datasets in both cases are the same as ours.

\begin{table}[!ht]
    \caption{\textbf{Po-Dec progressive exploration durations (environment steps) and residual scales.} We skip Po-Dec on \ib due to the high hardware and time cost of this task.}
    \label{tab:podec_schedule_scale}
    \centering
    \footnotesize
    \begin{tabular}{lcc}
        \toprule
        Task & Duration & Residual scale \\
        \midrule
        \texttt{RM\_can} & 200k & 0.10 \\
        \texttt{RM\_square} & 200k & 0.05 \\
        \texttt{RM\_transport} & 20k & 0.02 \\
        \texttt{Adroit\_door} & 200k & 0.15 \\
        \texttt{Adroit\_hammer} & 200k & 0.10 \\
        \texttt{Adroit\_pen} & 200k & 0.25 \\
        \texttt{MW\_push-wall} & 100k & 0.20 \\
        \texttt{MW\_box-close} & 100k & 0.30 \\
        \midrule
        \po & 16k & 0.02 \\
        \fe & 16k & 0.02 \\
        \ob & 16k & 0.02 \\
        \bottomrule
    \end{tabular}
    \vspace{-2mm}
\end{table}

\subsubsection{Real-World Platform}\label{app:platform}
The real-world experiments are conducted with an xArm-6 (in \po) and a Franka Emika Panda (in \fe), both with third-person-view RGB cameras and Robotiq 2F-85 grippers. In \ob and \ib, we use these two arms together, with one wrist-mounted camera on each arm in addition to the third-person-view camera.
Policies output desired end-effector pose commands (and gripper width if applicable) executed by a low-level interpolation controller. A 3D-printed spatula holder is used in \fe, and two 3D-printed rubber tong holders are used in \po.

\subsubsection{Real-World Runtime Profiling}\label{app:runtime}

We profile \po and \ib on the online training and deployment machine (RTX 4090D, i9-14900KF), covering higher control frequency and richer multi-camera inputs, respectively. Table~\ref{tab:latency} reports inference latency over 20 evaluation episodes, excluding the first prediction of each episode. Each prediction generates a full action chunk, which is then executed over 16 or 24 steps. Mean end-to-end latency is 28.07 and 37.62\,\si{ms}, below the respective control periods of 33.3 and 50\,\si{ms}. The VIB encoder, RL actor, and VIB decoder required by ZPRL together take 1.36 and 0.52\,\si{ms}, accounting for only 4.8\% and 1.4\% of the total. The observation encoder and iterative flow denoising dominate inference cost.

\begin{table}[!ht]
    \caption{\textbf{Per-module inference latency of ZPRL on two real-world tasks (\si{ms}).} Values are mean~$\pm$~standard deviation over all predictions of 20 evaluation episodes, excluding the first prediction of each run. Modules marked with $\dagger$ are the components added by ZPRL. The last row gives their combined share of the end-to-end latency.}
    \label{tab:latency}
    \centering
    \footnotesize
    \setlength{\tabcolsep}{4pt}
    \renewcommand{\arraystretch}{1.05}
    \begin{tabular}{@{}lcc@{}}
        \toprule
        \makecell[l]{Module} &
        \makecell[c]{\texttt{Place Orange}\\(30\,\si{Hz})} &
        \makecell[c]{\texttt{Insert Bills}\\(20\,\si{Hz})} \\
        \midrule
        Obs. encoder             & $6.38 \pm 1.88$ & $9.16 \pm 4.21$ \\
        VIB encoder $\dagger$    & $0.69 \pm 0.44$ & $0.28 \pm 0.27$ \\
        RL actor $\dagger$       & $0.38 \pm 0.32$ & $0.13 \pm 0.14$ \\
        VIB decoder $\dagger$    & $0.29 \pm 0.28$ & $0.11 \pm 0.12$ \\
        Flow gen. (one pass)     & $8.60 \pm 4.62$ & $2.42 \pm 1.24$ \\
        Flow gen. (all passes)   & $19.19 \pm 8.11$ & $27.15 \pm 5.07$ \\
        \textbf{End-to-end}      & $\mathbf{28.07 \pm 9.75}$ & $\mathbf{37.62 \pm 7.91}$ \\
        \midrule
        Share of $\dagger$ modules & 4.8\% & 1.4\% \\
        \bottomrule
    \end{tabular}
    \vspace{-2mm}
\end{table}

During training, the main process synchronously alternates between collecting episodes and updating models. Updates occur only between episodes, leaving controller execution uninterrupted. At batch size 256, an update takes $3.00\pm2.86$\,\si{ms} on \po and $5.33\pm3.90$\,\si{ms} on \ib, averaged over 1,000 updates, corresponding to 333.4 and 187.5 updates per second. These rates measure optimizer throughput, not robot control frequency.

\subsection{What Is ZPRL Doing?}\label{app:whatzprl}

\begin{figure*}[!ht]
    \centering
    \includegraphics[width=0.98\textwidth]{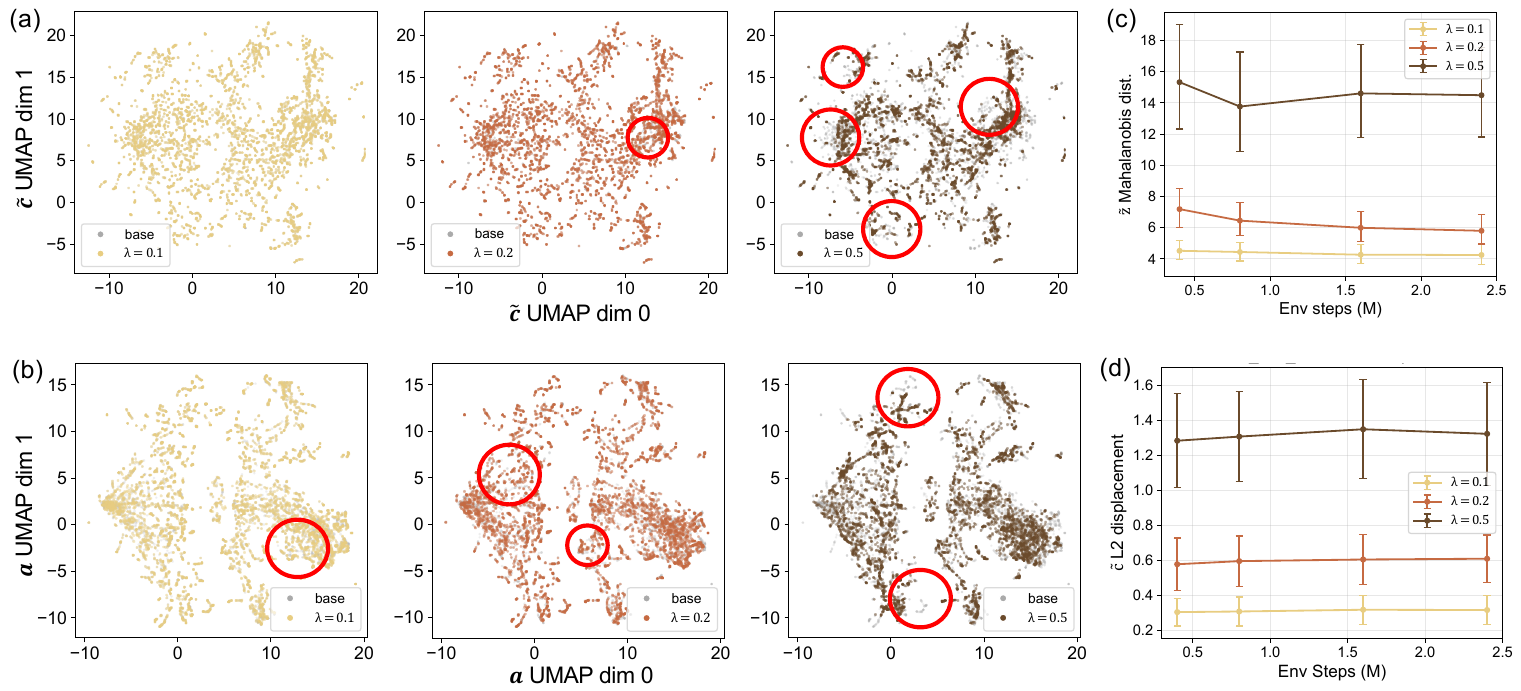}
    \vspace{-3mm}
    \caption{\textbf{What ZPRL changes during online finetuning.} (a) UMAP projections of the decoded observation embedding $\tilde{\vc}$ and (b) the generated action $\va$ on \texttt{Square} at 0.4M environment steps, comparing samples from the base policy and ZPRL policies under different perturbation scales $\lambda$. The SRs for the three intermediate checkpoints are 0.51 ($\lambda=0.1$), 0.56 ($\lambda=0.2$), 0.00 ($\lambda=0.5$), respectively. Red circles highlight representative regions with clear mismatch after steering. (c) Mean Mahalanobis distance of perturbed latents $\tilde{\vz}$ to the offline latent distribution and (d) mean L2 displacement between decoded embeddings with and without perturbation. Larger $\lambda$ induces stronger latent shifts and larger decoded-feature changes.}
    \label{fig:umap}
    \vspace{-3mm}
\end{figure*}

To better understand what ZPRL is doing, and especially how the perturbation scale $\lambda$ affects training, we visualize how online RL changes the policy's intermediate representations and outputs on the \texttt{Square} task. We use UMAP (Uniform Manifold Approximation and Projection~\cite{mcinnes2020umap}) to project the decoded observation embedding $\tilde{\vc}$ and the generated action $\va$, before and after online finetuning, to a 2D plane. We focus on checkpoints at 0.4M environment steps but similar patterns also appear at other stages of training.

As shown in \Figref{fig:umap}(a) and (b), ZPRL appears to mainly \emph{remap} state--action associations rather than invent entirely new actions. After adding the perturbation $\lambda \Delta \vz$ to the base latent, the RL policy changes the decoded feature that conditions the frozen action generator, so that the sampled action becomes better aligned with the reward objective given the current state instead of simply replaying behavior from the offline dataset~\cite{he2026demystifying}. In the UMAP visualizations, the point clouds of $\tilde{\vc}$ and $\va$ after finetuning still largely overlap with those of the base policy, but their local density and pairing structure are reorganized; representative mismatched regions are highlighted by red circles. A larger $\lambda$ leads to a stronger reorganization.

However, this steering must remain local to be effective. To quantify this, we track both the out-of-distribution (OOD) degree of the perturbed latent $\tilde{\vz}$ and the displacement of the decoded feature caused by the perturbation. Specifically, we fit a Gaussian distribution to 10,000 latent samples $\vz$ from the offline \texttt{Square} dataset using a Ledoit--Wolf covariance estimator~\cite{Ledoit2004well}, and compute the Mahalanobis distance~\cite{mahalanobis1936generalised} of perturbed latents $\tilde{\vz}$ produced by the online policy. We also compute the L2 distance between the decoded feature with perturbation, $\tilde{\vc}_{\mathrm{on}}$, and that without perturbation, $\tilde{\vc}_{\mathrm{off}}$. All computations are implemented with \texttt{sklearn}~\cite{scikit-learn}.

As shown in \Figref{fig:umap}(c) and (d), increasing $\lambda$ consistently increases both the OOD score of $\tilde{\vz}$ and the decoded-feature displacement. When $\lambda$ becomes too large, the frozen VIB decoder is forced to extrapolate to latents far outside its training support. The resulting decoded feature then deviates substantially from the original one, making the downstream action generator produce less meaningful actions for the current state. In our experiments, keeping the average L2 displacement between $\tilde{\vc}_{\mathrm{on}}$ and $\tilde{\vc}_{\mathrm{off}}$ on the offline dataset below about 0.8 works well on \texttt{Square}.

\subsection{Different \texorpdfstring{$\beta$}{beta} During Offline Training}\label{app:beta}

\begin{figure}[!ht]
    \centering
    \includegraphics[width=0.6\columnwidth]{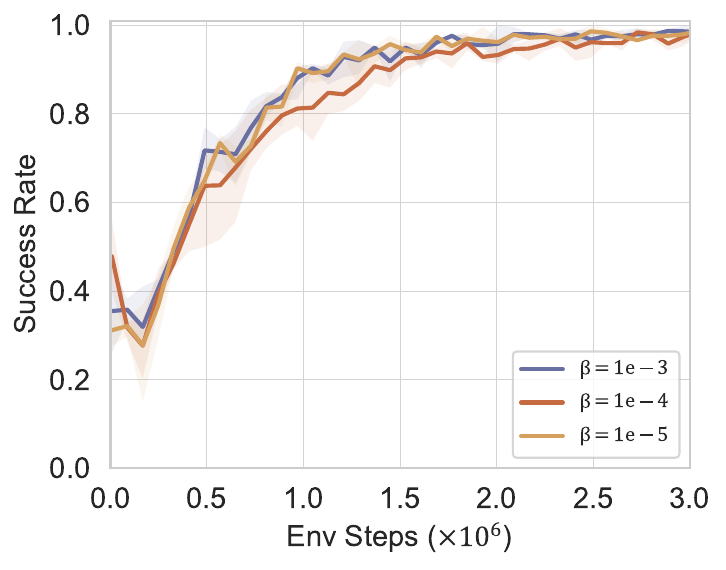}
    \vspace{-3mm}
    \caption{\textbf{Online RL finetuning on \texttt{Square} starting from base policies trained with different KL weights $\beta$.} All variants use the same online setting with $\lambda=0.2$. The three learning curves are highly similar, indicating limited sensitivity to $\beta$ within $\{10^{-5}, 10^{-4}, 10^{-3}\}$.}
    \vspace{-5mm}
    \label{fig:sim_beta}
\end{figure}

We also ablate the KL weight $\beta$ used in the VIB objective during offline training. Starting from base policies trained with different $\beta$ values, we run the same online RL finetuning procedure on \texttt{Square} with a fixed perturbation scale $\lambda=0.2$. As shown in \Figref{fig:sim_beta}, the resulting learning curves are highly similar across $\beta=10^{-5}, 10^{-4}, 10^{-3}$, with no clear difference in either convergence speed or final success rate. The RMS of per-dimension posterior standard deviation is approximately 0.002, 0.01, and 0.16 for $\beta=10^{-5},10^{-4},10^{-3}$, respectively. Thus, the KL term changes the posterior scale, while online adaptation remains robust across this two-order-of-magnitude range. We use it as a light regularizer with default $\beta=10^{-4}$, alongside the action-prediction objective, to prevent arbitrary latent scaling. This is because the perturbation scale $\lambda$ is defined relative to the latent magnitude. Small KL weights can also be found in latent diffusion models (e.g., $10^{-6}$ in~\cite{rombach2022ldm}).

\subsection{Deterministic versus Stochastic Bottlenecks}\label{app:deterministic}

\begin{figure}[!ht]
    \centering
    \includegraphics[width=0.49\columnwidth]{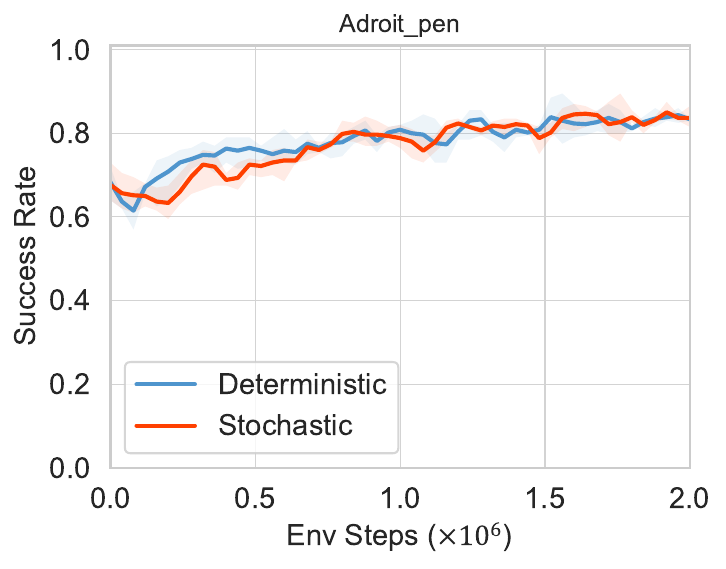}\hfill
    \includegraphics[width=0.49\columnwidth]{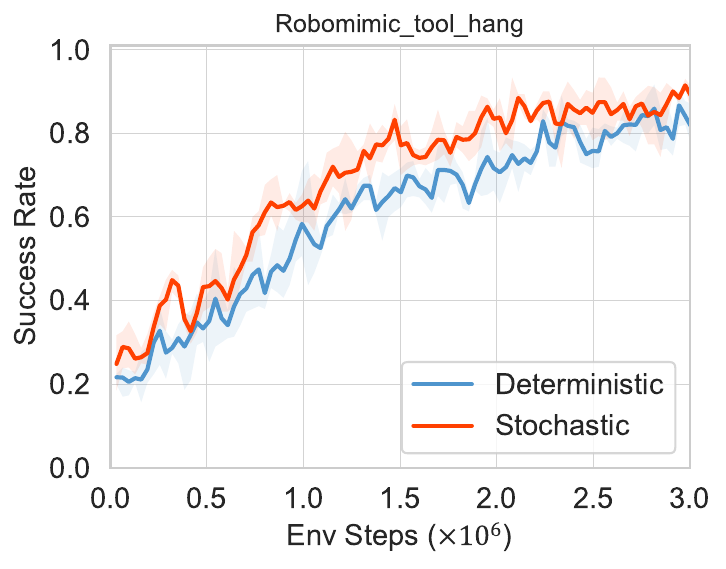}
    \vspace{-2mm}
    \caption{\textbf{RL learning curves with deterministic and stochastic latents.} Left: \texttt{Pen}. Right: \texttt{Tool\_hang}.}
    \label{fig:z_type}
    \vspace{-3mm}
\end{figure}

We replace the Gaussian VIB with an architecture-matched deterministic encoder-decoder that removes the variance head, sampling, and KL loss. Then we test both interfaces on \texttt{Adroit\_pen} and \texttt{Robomimic\_tool\_hang}. Both variants learn successfully. Their curves closely overlap on \texttt{Pen}, while the stochastic variant is slightly better through most of training on \texttt{Tool\_hang} (\Figref{fig:z_type}). These results suggest that the compact, task-aligned bottleneck is the common effective interface, while the benefit of stochastic regularization is task-dependent. We keep VIB as the default interface inherited from SOE~\cite{jin2025soe}, without claiming that variational stochasticity is universally necessary for ZPRL.

\subsection{Post-hoc Fitting of the Bottleneck Interface}\label{app:posthoc}

We evaluate whether the VIB interface can be attached after base-policy training on \texttt{Robomimic\_transport} and \texttt{Metaworld\_box-close}. We first train a standard flow policy without a VIB, then freeze its observation encoder and flow generator and fit only a newly attached VIB encoder-decoder. The VIB uses the frozen observation embeddings and paired actions, with the same flow-matching action-prediction objective and KL regularizer as in joint training. The jointly trained and post-hoc variants subsequently use the same online RL procedure, with both the base policy and VIB frozen.

\begin{figure}[!ht]
    \centering
    \includegraphics[width=0.49\columnwidth]{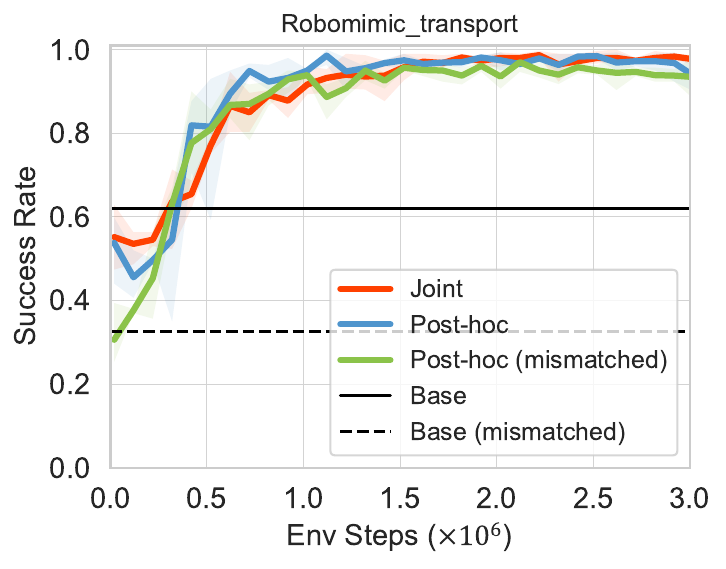}\hfill
    \includegraphics[width=0.49\columnwidth]{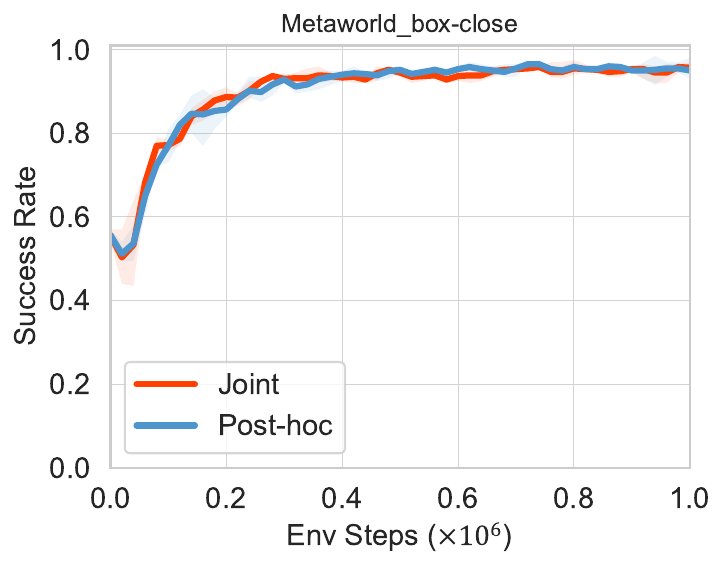}
    \vspace{-2mm}
    \caption{\textbf{Online RL with jointly and post-hoc trained VIB.} Left: \texttt{Transport}, including post-hoc fitting on PH data for an MH-trained base. Right: \texttt{Box-close}.}
    \label{fig:posthoc}
    \vspace{-2mm}
\end{figure}

As shown in \Figref{fig:posthoc}, post-hoc and joint training yield closely matched online performance on both tasks, despite minor early-stage differences on \texttt{Transport}. Thus, the VIB interface need not be optimized concurrently with the base policy, allowing ZPRL to steer an independently trained checkpoint without retraining its backbone.

We also test attachment without the original dataset. On \texttt{Transport}, we fit the post-hoc VIB on proficient-human (PH) data while the frozen base was trained on multi-human (MH) data. This mismatched VIB path starts from a lower SR than the matched variants, but online RL largely closes the gap. Its learning curve tracks the matched variants and converges only a few percentage points below them. PH and MH cover the same task and scenes but differ mainly in operator proficiency. These results support post-hoc fitting with task-relevant replacement data under moderate distribution shift. Only offline observation-action pairs (or encoded features) and an accessible intermediate interface are needed. Arbitrary cross-task transfer is an open question for future work.

\subsection{Cross-task Ablations}\label{app:cross_task}

\begin{figure}[!ht]
    \centering
    \includegraphics[width=0.49\columnwidth]{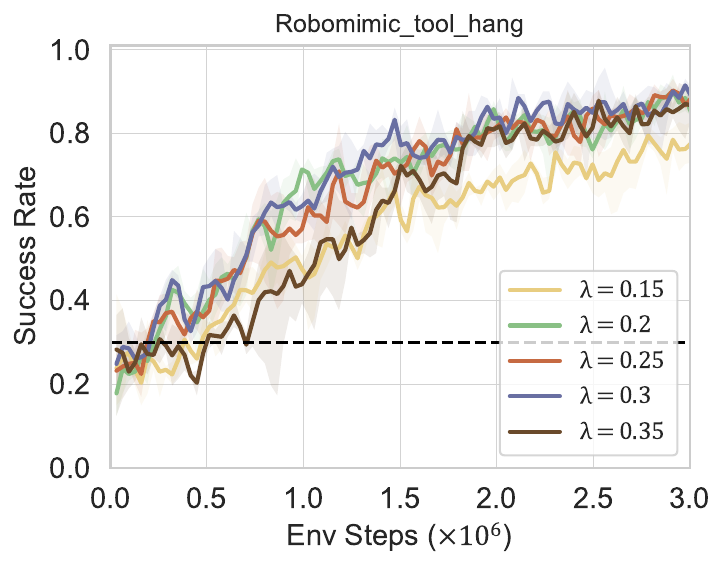}\hfill
    \includegraphics[width=0.49\columnwidth]{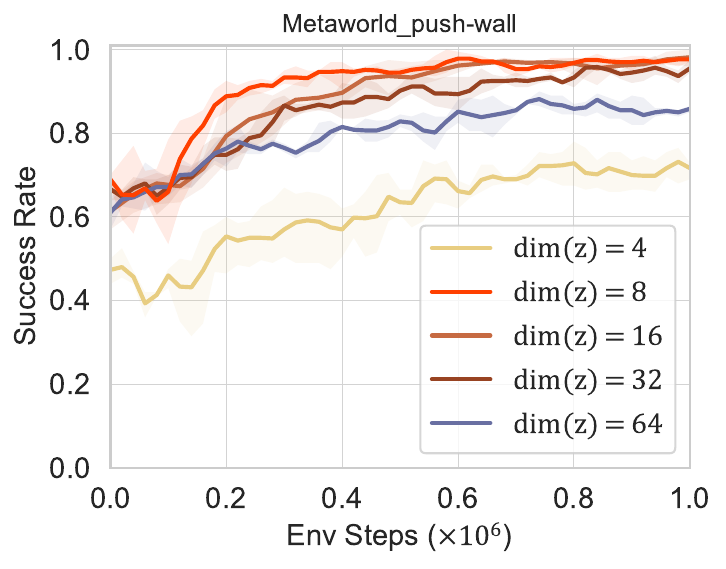}\\
    \includegraphics[width=0.49\columnwidth]{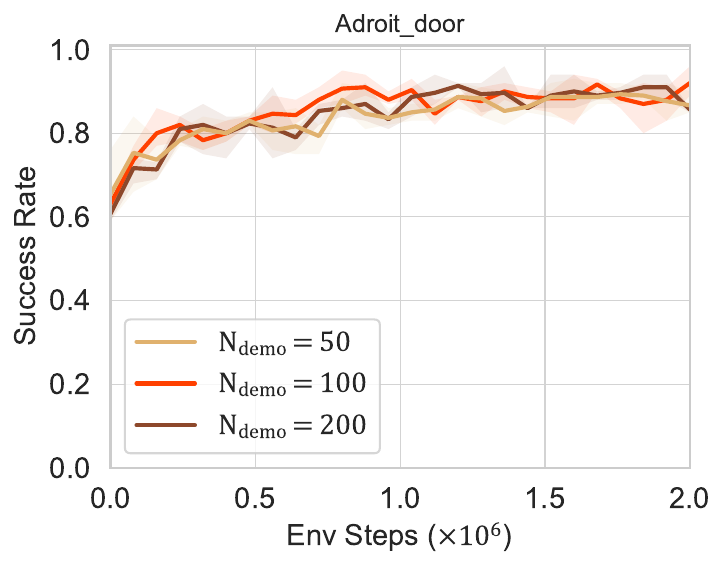}
    \vspace{-2mm}
    \caption{\textbf{Cross-task ablations.} Perturbation scale on \texttt{Tool\_hang} (top left), latent dimension on \texttt{Push-wall} (top right), and offline dataset size on \texttt{Door} (bottom).}
    \label{fig:more_ablation}
    \vspace{-2mm}
\end{figure}

We extend key ablations beyond \texttt{Square}: perturbation scale on \texttt{Tool\_hang}, latent dimension on \texttt{Push-wall}, and offline dataset size on \texttt{Door} (\Figref{fig:more_ablation}). The scale and dimension trends hold: moderate perturbations support effective steering, and $\dim(\vz)=8$ performs best over most of training on \texttt{Push-wall}, while further compression reduces performance and larger latents provide no benefit.

The dataset-size result differs from \texttt{Square}: all variants perform similarly on \texttt{Door}. A possible explanation is that its data come from a medium-expert RL checkpoint and are more homogeneous than the multi-human data in \texttt{Robomimic}, so additional trajectories may add little behavioral coverage. The benefit of increasing offline data is therefore task- and data-dependent.

\subsection{Interaction Efficiency across Simulation Tasks}\label{app:interaction}

Table~\ref{tab:train_stats} compares DSRL, Po-Dec, and ZPRL across all eight tasks. Threshold-crossing steps are averaged only over seeds that reach the specified SR; N/A indicates that no seed reaches it. Final SR is averaged over the last 5 data points and across seeds. ZPRL reaches high SR earlier than Po-Dec wherever both methods reach the threshold, despite their comparable final performance.

\begin{table}[!ht]
    \caption{\textbf{Training-time RL efficiency and final performance across the eight simulation tasks.} We report the environment steps (in millions, averaged over seeds that reach each threshold) to first reach SR$=$0.9 and 0.95, and the final SR averaged over the last 5 points. N/A means that none of the seeded runs reaches the threshold. The best value of each task and metric is in \textbf{bold}.}
    \label{tab:train_stats}
    \par\vspace*{2mm}
    \centering
    \footnotesize
    \setlength{\tabcolsep}{2pt}
    \renewcommand{\arraystretch}{1.05}
    \begin{tabular}{@{}llccc@{}}
        \toprule
        \makecell[l]{Task} &
        \makecell[l]{Method} &
        \makecell[c]{Steps (M)\\to SR$=$0.9 $\downarrow$} &
        \makecell[c]{Steps (M)\\to SR$=$0.95 $\downarrow$} &
        \makecell[c]{Final\\SR $\uparrow$} \\
        \midrule

        \multirow{3}{*}{\texttt{RM\_can}}
        & DSRL   & 0.16 & 0.20 & \textbf{0.99} \\
        & Po-Dec & 0.13 & 0.16 & 0.95 \\
        & ZPRL   & \textbf{0.08} & \textbf{0.13} & 0.98 \\
        \midrule

        \multirow{3}{*}{\texttt{RM\_square}}
        & DSRL   & 1.17 & 1.88 & 0.95 \\
        & Po-Dec & 1.68 & 2.22 & 0.92 \\
        & ZPRL   & \textbf{1.15} & \textbf{1.55} & \textbf{0.98} \\
        \midrule

        \multirow{3}{*}{\texttt{RM\_transport}}
        & DSRL   & 1.58 & 2.01 & 0.96 \\
        & Po-Dec & 0.89 & 1.14 & 0.96 \\
        & ZPRL   & \textbf{0.66} & \textbf{1.10} & \textbf{0.98} \\
        \midrule

        \multirow{3}{*}{\texttt{Adroit\_door}}
        & DSRL   & N/A  & N/A & 0.69 \\
        & Po-Dec & 1.10 & N/A & 0.85 \\
        & ZPRL   & \textbf{0.87} & \textbf{1.56} & \textbf{0.91} \\
        \midrule

        \multirow{3}{*}{\texttt{Adroit\_hammer}}
        & DSRL   & \textbf{0.08} & 0.39 & \textbf{0.95} \\
        & Po-Dec & 0.31 & 0.52 & 0.91 \\
        & ZPRL   & \textbf{0.08} & \textbf{0.34} & 0.93 \\
        \midrule

        \multirow{3}{*}{\texttt{Adroit\_pen}}
        & DSRL   & N/A & N/A & 0.76 \\
        & Po-Dec & N/A & N/A & 0.74 \\
        & ZPRL   & \textbf{1.60} & N/A & \textbf{0.82} \\
        \midrule

        \multirow{3}{*}{\texttt{MW\_box-close}}
        & DSRL   & N/A  & N/A  & 0.66 \\
        & Po-Dec & 0.25 & 0.53 & \textbf{0.94} \\
        & ZPRL   & \textbf{0.15} & \textbf{0.20} & 0.93 \\
        \midrule

        \multirow{3}{*}{\texttt{MW\_push-wall}}
        & DSRL   & 0.60 & N/A  & 0.78 \\
        & Po-Dec & 0.25 & 0.31 & 0.96 \\
        & ZPRL   & \textbf{0.17} & \textbf{0.29} & \textbf{0.98} \\

        \bottomrule
    \end{tabular}
\end{table}

\subsection{Comparison with SOE}\label{app:soe}

We compare ZPRL with SOE to examine the benefit of reward-driven latent steering over iteratively imitating successful rollouts (demonstrations included). We evaluate both methods on \texttt{MetaWorld\_box-close} and \texttt{Adroit\_hammer} under a budget of 1,000 interaction episodes.

\begin{figure}[!ht]
    \centering
    \includegraphics[width=0.49\columnwidth]{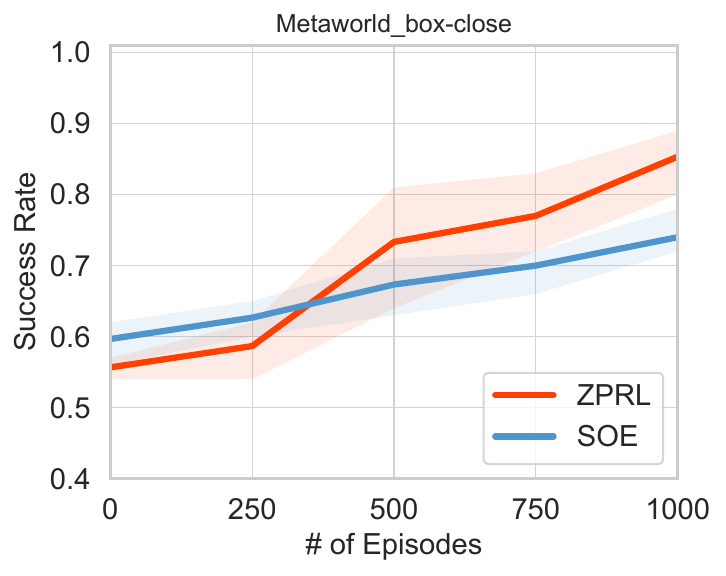}\hfill
    \includegraphics[width=0.49\columnwidth]{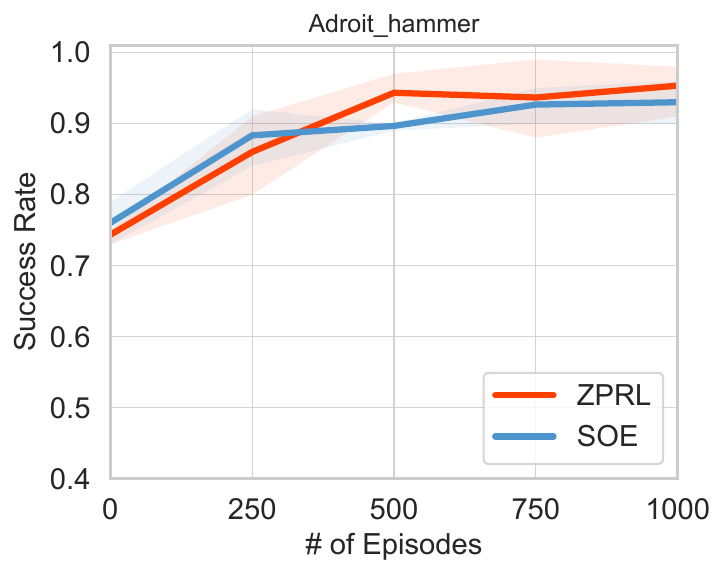}
    \caption{\textbf{Comparison with SOE under matched interaction budgets} on \texttt{Box-close} (left) and \texttt{Hammer} (right). SOE performs four rounds of 250 exploratory episodes and 50k gradient steps of retraining per round. ZPRL is evaluated at the corresponding episode counts. Each checkpoint is evaluated on 100 trials per seed, with the VIB disabled for SOE and stochastic sampling disabled for ZPRL.}
    \label{fig:soe_comparison}
    \vspace{-2mm}
\end{figure}

Starting from the offline-trained base policies, SOE alternates between exploration and retraining for four rounds. Each round collects 250 episodes by sampling the latent from the Gaussian distribution given by the VIB encoder (doubling $\vz$'s standard deviation for exploration, as suggested by SOE's official implementation), and retrains the flow policy and its VIB from scratch for 50,000 gradient steps on the cumulative successful rollouts. The growing dataset also includes the original demonstrations.
ZPRL instead updates its actor and critic while keeping the base policy frozen. We evaluate at 0, 250, 500, 750, and 1,000 collected episodes. The final budget corresponds to about 120k and 200k environment steps for ZPRL on \texttt{Box-close} and \texttt{Hammer}, respectively.
For this comparison, both methods use 5 flow steps. ZPRL alternates data collection and policy updates every 500 action chunks instead of 1,000 on \texttt{Hammer} to track the episode counts more precisely. These changes apply only to this comparison, and all other experiments keep their original protocol. The results thus differ slightly from those in the main text. Each evaluation uses 100 trials per seed over three seeds. SOE uses its base path bypassing the VIB module, and ZPRL is evaluated deterministically.

Both methods improve over their initial performance (\Figref{fig:soe_comparison}). SOE leads at 250 episodes but ZPRL overtakes it by 500 episodes and achieves higher SRs at 1,000 episodes on both tasks.
Although SOE uses task success to filter data, it aims at optimizing an imitation objective. In contrast, ZPRL optimizes the steering policy for task return. Such a trend of limited gains from successful-episode imitation is also reported and analyzed in \cite{mirchandani2024so}. These results show that reward-driven steering can achieve higher task success than iterative self-imitation given the same and sufficient interaction budget.

\subsection{Real-World Smoothness Evaluation}\label{app:real_smooth}

We evaluate the final Po-Dec and ZPRL checkpoints on \po with 20 rollouts per method, using deterministic actions and no temporal filter. All trajectories are included regardless of outcome, with episodes terminated upon success, irrecoverable failure, or timeout. We compute translational velocity, acceleration, and jerk from the logged end-effector (EE) positions by finite differences without smoothing. Table~\ref{tab:real_smooth_stats} reports the mean and sample standard deviation of their L2 norms at two levels: pooled over all control steps, and over the 20 per-trajectory means.

The pooled standard deviations exceed their means because differentiating logs twice and three times without smoothing amplifies per-step noise. Averaging within each trajectory removes most of this dispersion, and the two methods separate cleanly at that level. ZPRL halves acceleration (1.8 to 0.9\,\si{m/s^2}) and reduces jerk by 47\% (87.8 to 46.3\,\si{m/s^3}) at similar mean velocity. ZPRL therefore executes the task with consistently smoother motion.

\begin{table}[!ht]
    \caption{\textbf{Real-world smoothness on \po.} Velocity, acceleration, and jerk are obtained from the logged EE positions by finite differences without smoothing. Per-step statistics pool all control steps of the 20 rollouts per method (roughly 5,000 samples each), while per-trajectory statistics are taken over the 20 per-rollout means.}
    \label{tab:real_smooth_stats}
    \centering
    \footnotesize
    \setlength{\tabcolsep}{4pt}
    \renewcommand{\arraystretch}{1.05}
    \begin{tabular}{@{}lccc@{}}
        \toprule
        \makecell[l]{Method} &
        \makecell[c]{Vel. (\si{m/s})} &
        \makecell[c]{Acc. (\si{m/s^2}) $\downarrow$} &
        \makecell[c]{Jerk (\si{m/s^3}) $\downarrow$} \\
        \midrule
        \multicolumn{4}{@{}l}{\emph{Per control step (pooled)}} \\
        Po-Dec & $0.09 \pm 0.07$ & $1.8 \pm 2.6$ & $88.0 \pm 132.1$ \\
        \textbf{ZPRL} & $0.09 \pm 0.05$ & $0.9 \pm 1.2$ & $46.3 \pm 71.4$ \\
        \midrule
        \multicolumn{4}{@{}l}{\emph{Per trajectory ($n=20$)}} \\
        Po-Dec & $0.09 \pm 0.006$ & $1.8 \pm 0.4$ & $87.8 \pm 24.9$ \\
        \textbf{ZPRL} & $0.09 \pm 0.004$ & $0.9 \pm 0.2$ & $46.3 \pm 12.2$ \\
        \bottomrule
    \end{tabular}
    \vspace{-2mm}
\end{table}